\documentclass[conference]{sty/IEEEtran}

\IEEEoverridecommandlockouts

\usepackage{amsmath,amsfonts,amssymb}
\usepackage{algorithmic}
\usepackage{algorithm}
\usepackage{array}
\usepackage[caption=false,font=normalsize,labelfont=sf,textfont=sf]{subfig}
\usepackage{textcomp}
\usepackage{stfloats}
\usepackage{url}
\usepackage{verbatim}
\usepackage{booktabs}
\usepackage{multirow}
\usepackage{rotating}
\usepackage{hyperref}
\usepackage{makecell}
\usepackage{cite}
\usepackage{xspace}
\usepackage{sty/my_acronyms}
\usepackage{siunitx}
\usepackage{flushend}
\usepackage{tikz}
\usetikzlibrary{shapes,arrows,calc, intersections}
\newcommand{\RomanNumeralCaps}[1] {\MakeUppercase{\romannumeral #1}}

\newcommand{\magyc}{\mbox{MAGYC}\xspace}

\newcommand{\magycbfg}{\mbox{MAGYC-BFG}\xspace}
\newcommand{\magycifg}{\mbox{MAGYC-IFG}\xspace}
\newcommand{\twostep}{\mbox{TWOSTEP}\xspace}
\newcommand{\magfactor}{\mbox{MagFactor3}\xspace}

\DeclareSIUnit\milligauss{mG}

\usepackage{marginnote}
\makeatletter
\newcommand{\smarginnote}[1]{\begingroup\if@firstcolumn\reversemarginpar\fi\marginnote{#1}\normalmarginpar\endgroup}
\makeatother
\newcommand{\tfflag}[1]{\smarginnote{\color{blue}\scriptsize{\fbox{#1}}}}
\newcommand{\tfflagl}[1]{\reversemarginpar\marginnote{\color{blue}\scriptsize{\fbox{#1}}}\normalmarginpar}
\newcommand{\tfflagr}[1]{\normalmarginpar\marginnote{\color{blue}\scriptsize{\fbox{#1}}}}

\renewcommand{\tfflag}[1]{}
\renewcommand{\tfflagl}[1]{}
\renewcommand{\tfflagr}[1]{}

\begin{document}

\title{Full Magnetometer and Gyroscope Bias Estimation using Angular Rates: Theory and Experimental Evaluation of a Factor Graph-Based Approach

% Funding-related thanks
\thanks{This work was supported by the David and Lucile Packard Foundation and FONDECYT-Chile under grant 11180907.}
}

% Authors
\author{\IEEEauthorblockN{Sebastián Rodríguez-Martínez}
\IEEEauthorblockA{\textit{Monterey Bay Aquarium Research Institute} \\
Moss Landing, California 95039--9644\\
srodriguez@mbari.org}
\and
\IEEEauthorblockN{Giancarlo Troni}
\IEEEauthorblockA{\textit{Monterey Bay Aquarium Research Institute} \\
Moss Landing, California 95039--9644\\
gtroni@mbari.org}
}

\maketitle

\begin{abstract}

Despite their widespread use in determining system attitude, \acp{mems} \acl{ahrs} are limited by sensor measurement biases. This paper introduces a method called MAgnetometer and GYroscope Calibration (\magyc), leveraging three-axis angular rate measurements from an angular rate gyroscope to estimate both the hard- and soft-iron biases of magnetometers as well as the bias of gyroscopes. We present two implementation methods of this approach based on batch and online incremental factor graphs. Our method imposes fewer restrictions on instrument movements required for calibration, eliminates the need for knowledge of the local magnetic field magnitude or instrument's attitude, and facilitates integration into factor graph algorithms for Smoothing and Mapping frameworks. We validate the proposed methods through numerical simulations and in-field experimental evaluations with a sensor onboard an underwater vehicle. By implementing the proposed method in field data of a seafloor mapping dive, the \acl{dr}-based position estimation error of the underwater vehicle was reduced from \SI{10}{\percent} to \SI{0.5}{\percent} of the \acl{dt}. For additional material and information, visit the \magyc project page: \href{https://compas-docs.github.io/magyc/}{https://compas-docs.github.io/magyc/}.

\end{abstract}


% reset acronyms
\acresetall

%%%%%%%%%%%%%%%%%%%%%%%%%%%%%%%%%%%%%%%%%%%%%%%%%%%%%%%%%%%%%%%%%%%%%%%%%%%%%%%%
% Introductions
\section{Introduction}
\label{sec:intro}

%%%%%%%%%%%%%%%%%%%%%%%%%%%%%%%%%%%%
% General Intro
%
\noindent \ac{mems} \acp{ahrs} find wide applications across various domains for determining system attitude, particularly in vehicle navigation systems operating in space, ground, and marine environments. They play a vital role in GPS-denied applications, such as deployments of \acp{auv} or navigation in confined spaces like mines and tunnels, where inertial navigation serves as the primary method for estimating system pose. Despite their significance, these sensors are prone to calibration errors and often yield imprecise navigation results. While expensive high-grade sensors ($>\!50k$ USD) can partially mitigate this issue, addressing calibration errors remains a challenge for mid and low-range sensors designed for widespread use, typically costing thousands of dollars.

A typical \ac{mems} \ac{ahrs} consists of a three-axis magnetometer, a three-axis accelerometer, a three-axis gyroscope, and a temperature sensor. The magnetometer measures the Earth's magnetic field locally, assisting in determining the system's heading. The accelerometer, under minimal external acceleration, measures the system's inclination relative to local gravity, providing pitch and roll orientation information. Lastly, the gyroscope provides the vehicle's angular rate, contributing to the refinement of rotation estimation.

%%%%%%%%%%%%%%%%%%%%%%%%%%%%%%%%%%%%
% AUV trajectories Diagrams
% Description: AUV showing three trajectories: RAW, MAGYC, and Ground Truth
%
\begin{figure}[t!]
\centering
\begin{tikzpicture}[scale=0.9]
\node[anchor=center,inner sep=0, rotate=33, anchor=center] at (-2.1,-0.03) {\includegraphics[width=3.5cm,height=1.3cm]{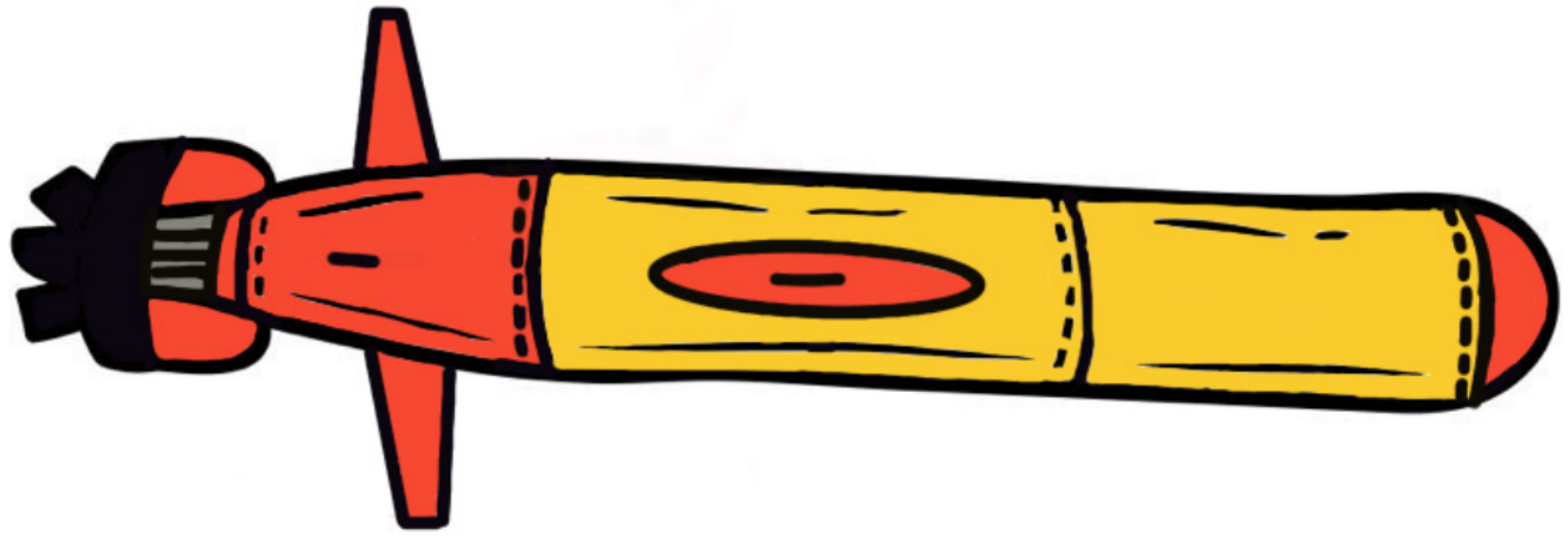}};
\draw [dashed, gray, ultra thick, rotate around={32:(-2,0)}] (-2,0) -- (5,0);
\node[rotate=32, gray] at (2.8,3.3) {Ground Truth};
\draw [dashed, black, ultra thick, rotate around={31:(-2,0)}] (-2,0) -- (5,0);
\node[rotate=31] at (3.2,2.85) {\magyc};
\draw [dashed, red, ultra thick, rotate around={20:(-2,0)}] (-2,0) -- (5,0);
\node[rotate=20, red] at (4.3,2.05) {Raw};
\draw [-, black, ultra thick] (-2,1.2) arc (90:32:1.2);
\draw [->, red, ultra thick] (-2,0) -- (-2,1.8) node[above] {\(\vec{m}\)};
\draw [->, black, ultra thick, rotate around={32:(-2,0)}] (-2,0) -- (0.5,0) node[below] {\(x\) };
\draw [->, black, ultra thick, rotate around={32:(-2,0)}] (-2,0) -- (-2,-1) node[below] {\(y\)};
\node[font=\small] at (-1.2, 1.3) {$\psi$};
\node at (0, -1.6) {Top View};
\end{tikzpicture}
\caption{Diagram illustrating an underwater vehicle's dead reckoning position using different magnetic field sources for the heading estimation with the corresponding trajectories represented with dashed lines.}
\label{fig:attitude_diagram}
\end{figure}
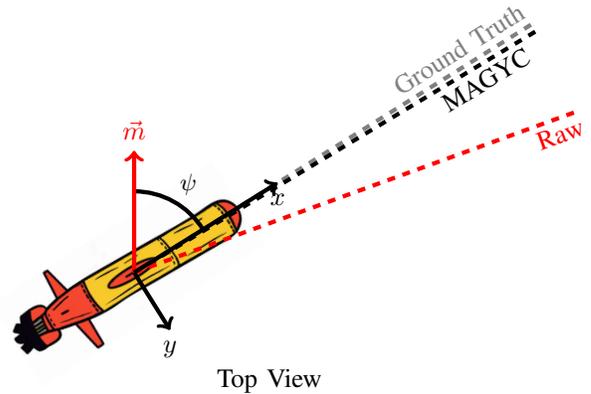

The accurate estimation of \ac{ahrs} components depends on effectively mitigating biases, scale factors, and non-orthogonality issues. Gyroscopes and accelerometers encounter challenges from biases, while magnetometers are vulnerable to calibration errors induced by nearby ferrous materials or electric currents, which can distort the magnetic field, leading to inaccuracies in heading estimation. There are two primary types of magnetometer calibration errors: hard-iron biases, caused by permanent magnetic fields from the vehicle and onboard instruments, resulting in constant output bias, and soft-iron biases, arising from magnetic materials near the sensor, which distort the magnetic field with scale factors and non-orthogonalities.

%%%%%%%%%%%%%%%%%%%%%%%%%%%%%%%%%%%%
% Problem: Related Work
% 
Accurate calibration of three-axis magnetometers is crucial for reliable attitude estimation, and several methods have been proposed to estimate calibration parameters without requiring additional reference sensors. One widely adopted approach Alonso and Shuster introduced is the \twostep method. It initially estimates the magnetometer's bias \cite{Alonso2002a}, followed by estimating the scale and non-orthogonality factors as well using an iterative least squares minimization \cite{Alonso2002b}. The full calibration of a magnetometer can also be formulated as an ellipsoid fitting problem, which Vasconcelos et al. \cite{Vasconcelos2011} solved using a maximum likelihood estimate method. Several least squares methods have been reported as well \cite{Fang2011, Dorveaux2009, Batista2023}. However, these methods are impractical in field applications as the device must perform wide ranges of angular motion in all three degrees of freedom. This is infeasible for many devices mounted on full-scale vehicles, such as pitch- and roll-stable \acp{rov}. Furthermore, accurately knowing the Earth's local magnetic field is necessary for improved performance, which magnetic field models can calculate \cite{NOAA2023}, but it can present significant errors due to unmodeled local perturbations.

% 
% Reviewers' Reply
\tfflagr{R2-2}
Given the availability of inertial sensors with magnetometers as a package, some approaches have fused accelerometer and magnetometer measurements to estimate the magnetometer sensor biases, taking advantage of the accelerometer to measure the local gravity vector \cite{Kok2016, Papafotis2019, Ammann2015}. However, these methods are affected by errors in the calibration caused by translational accelerations of the system or require predefined calibration patterns. Troni and Whitcomb \cite{Troni2020} proposed a novel method using angular velocity measurements, which is limited to the magnetometer's hard-iron calibration, neglecting the soft-iron matrix effects and assuming that the angular velocity sensor is already bias-compensated. This method was later extended by Spielvogel and Whitcomb \cite{Spielvogel2018} to include the estimation of gyroscope and accelerometer biases, still neglecting the effects of the soft-iron calibration.

% 
% Reviewers' Reply
\tfflagr{R2-2}
The previously discussed methods are limited to batch calibrations, where all the measurements must be collected beforehand. However, it is desirable to perform calibration in the field when pre-calibration is not possible, for instance, due to changes in the vehicle's configuration deployed in highly disturbed environments. Crassidis et al. \cite{crassidis2005} presented an extension to the \twostep method that incorporates an Extended Kalman Filter (EKF) and an Unscented Kalman Filter (UKF). Later, Ma and Jing \cite{Ma2005}, and Soken and Sakai \cite{Soken2019} proposed alternative approaches for the UKF method, while Guo et al. \cite{Guo2008} presented an alternative EKF method. Additionally, Han et al. \cite{Han2017} and Spielvogel et al. \cite{Spielvogel2022} proposed a gyroscope-aided EKF, with the latter also incorporating gyroscope biases in the estimation. A different approach proposed by Lathrop et al. \cite{Lathrop2023} poses magnetometer's hard-iron calibration based on a factor graph optimization; however, this approach neglects the soft-iron and gyroscope bias and requires a predefined calibration routine. The presented approaches have in common a demand for substantial angular motion for accurate calibration and may experience difficulties in accurately estimating the actual values if the angular motion is insufficient.

%%%%%%%%%%%%%%%%%%%%%%%%%%%%%%%%%%%%
% Barriers/Advantages + Proposed Solution and contributions
% 
The previously reported approaches exhibit at least one of the following limitations that restrict their implementation scenarios: ($i$) the necessity for extensive angular motion in all three rotational degrees of freedom, often impractical for devices mounted on full-scale vehicles like pitch and roll-stable \acp{rov}; ($ii$) the requirement for knowledge of the local Earth's magnetic field magnitude from magnetic field models, which may introduce significant errors due to unmodeled local perturbations \cite{NOAA2023}; or ($iii$) the inability to determine the magnetometer's soft-iron error, neglecting non-orthogonalities and scale factors that affect systems surrounded by ferromagnetic materials such as iron and steel, resulting in less accurate output.

% 
% Reviewers' Reply
\tfflagr{R2-3}
This paper presents a novel approach called MAgnetometer and GYroscope Calibration (\magyc), effectively overcoming these challenges. The approach facilitates two factor graph-based methods. The \magyc approach offers several notable advantages, including ($i$) fewer restrictions on angular movement requirements, ($ii$) no requirement for information about the local magnetic field or system's attitude, ($iii$) complete calibration for both magnetometer and gyroscope and ($iv$) integration into factor graph algorithms for smoothing and mapping frameworks with tools such as GTSAM \cite{BORGLab2023} for problems such as \ac{slam}.

% 
% Reviewers' Reply
\tfflagr{R1-3}
This paper builds on the prior work of the authors, extending the foundational approach introduced in \cite{Rodriguez2024}, where a factor graph framework was proposed to estimate the full calibration of a three-axis magnetometer (including hard-iron and soft-iron effects) and a three-axis gyroscope using magnetometer and angular rate measurements. The primary advancement over the previous approach is the incorporation of system constraints directly into the model residual, leveraging the mathematical properties of the calibration parameters. This refinement enhances convergence and, consequently, improves post-calibration performance. Additionally, by applying the method to field data from mapping surveys, this work expands the evaluation to real-world navigation outcomes, demonstrating the robustness and applicability of the proposed method in operational settings.

%%%%%%%%%%%%%%%%%%%%%%%%%%%%%%%%%%%%
% Paper Structure
The paper is structured as follows. Section \RomanNumeralCaps{2} briefly overviews the mathematical content. Section \RomanNumeralCaps{3} details the sensors model, system model, and the proposed \magyc methods. Section \RomanNumeralCaps{4} describes the evaluation methodology, while Section \RomanNumeralCaps{5} reports simulation results, and Section \RomanNumeralCaps{6} presents experimental results. The results and conclusions are summarized in Section \RomanNumeralCaps{7}.

%%%%%%%%%%%%%%%%%%%%%%%%%%%%%%%%%%%%%%%%%%%%%%%%%%%%%%%%%%%%%%%%%%%%%%%%%%%%%%%%
% Mathematical Background
\section{Mathematical Background}
\label{sec:math}

%%%%%%%%%%%%%%%%%%%%%%%%%%%%%%%%%%%%
% Rotation Matrix
\subsection{Rotation Matrix}

\noindent The orientation of an instrument frame $V$ relative to an inertial world frame $W$ can be represented by a rotation matrix $\mathbf{R}(t) \in SO(3)$ \cite[Def. 3.1]{Lynch2017}.

%%%%%%%%%%%%%%%%%%%%%%%%%%%%%%%%%%%%
% Positive definite matrices
%
\subsection{Real Positive Definite Symmetric Matrices}
% 
% Reviewers' Reply
\tfflagr{R1-3}

\noindent A symmetric matrix is considered positive definite (PDS) if all its eigenvalues are positive. The set of all PDS matrices, denoted by $\mathcal{S}^n_{++}$, forms a cone manifold equipped with a natural Riemannian metric tensor. The geodesic distance between two matrices $\mathbf{A}$ and $\mathbf{B}$ in this space is given by the length of the shortest path connecting them \cite[Eq. 2.5]{Bonnabel2010}:

\begin{equation}
    \delta_2(\mathbf{A}, \mathbf{B}) = ||\log{(\mathbf{A}^{-\frac{1}{2}}\mathbf{B}\mathbf{A}^{-\frac{1}{2}}})||_F.
    \label{eq: geodesic_distance}
\end{equation}

%%%%%%%%%%%%%%%%%%%%%%%%%%%%%%%%%%%%
% Kronecker Product
%
\subsection{Kronecker Product}

\noindent The Kronecker product of the matrix $\mathbf{A} \in \mathbb{M}^{p \times q}$ with the matrix $\mathbf{B} \in \mathbb{M}^{r \times s}$ is defined as an $p \times q$ block matrix whose $(i, j)$ block is the $r \times s$ matrix $a_{ij}\mathbf{B}$ \cite[Def. 2.1]{Schacke2013}.

%%%%%%%%%%%%%%%%%%%%%%%%%%%%%%%%%%%%
% Operators
%
\subsection{Operators}

\subsubsection{Skew-Symmetric}

\noindent We denote $\left[ \; \cdot \; \right]_\times: \mathbb{R}^3 \rightarrow \mathbb{R}^{3\times3}$ as the usual skew-symmetric operator \cite[Def. 3.7]{Lynch2017}.

\subsubsection{vec-operator}

\noindent The vec-operator for any matrix $\mathbf{A} \in \mathbb{M}^{m \times n}$ is defined as the entries of $\mathbf{A}$ stacked columnwise, forming a vector of length $m \times n$ \cite[Def. 2.2]{Schacke2013}.

%%%%%%%%%%%%%%%%%%%%%%%%%%%%%%%%%%%%
% Factor Graphs
%
\subsection{Factor Graphs}

\noindent Factor graphs are representations of probabilistic or graphical models consisting of nodes representing unknown random variables ($x_i \in \mathcal {X} $) and edges representing the dependencies or relationships between these variables. The edges are associated with factors ($f_i \in \mathcal{F}$), which are probabilistic constraints derived from measurements, prior knowledge, or relationships between variables, and can be categorized as unary factors when connecting to a single node or binary factors if they connect two or more nodes \cite{Dellaert2012, Koller2009, Kschischang2001}.

The factor graph is a concise and intuitive way of representing a model that is well-suited for performing various types of probabilistic inference tasks, such as bias estimation given a set of measurements for sensor calibration. In comparison to traditional filtering methods, factor graphs have shown superior capability in managing nonlinear processes and measurement models. They offer notable advantages in processing time, flexibility, and modularity, particularly for large-scale and complex optimization problems \cite{Dellaert2006, Dai2022}.

Unary and binary factors can be represented as measurement likelihood with a Gaussian noise model \cite{Dellaert2006}, which is only evaluated as a function of the unknown state $\mathbf{x}$ since the measurement $m$ is considered known

\begin{equation}
L(\mathbf{x}; m) = \exp \left\{ -\frac{1}{2} ||h(\mathbf{x}) - m||_\mathbf{\Sigma}^2 \right\},
\label{eq: unary_factor}
\end{equation}

\noindent where $h(\mathbf{x})$ is a nonlinear measurement model, $|| \, \cdot \, ||$ is the Mahalanobis distance, and $\mathbf{\Sigma}$ is the noise covariance matrix.

Utilizing the factors and a prior belief over the unknown variable $P(\mathbf{x_0})$, we formulate the joint probability model. Our objective is to determine the optimal set of parameters $\mathbf{X}^*$ through the maximum a posteriori (MAP) estimate by maximizing the joint probability $P(\mathbf{X}, \mathbf{Z})$, leading to a nonlinear least squares problem (\ref{eq: factor_map}). Leveraging the factor graph structure and sparse connections in this estimation process helps reduce computational complexity \cite{Dellaert2006, Dellaert2012}.

\begin{equation}
\mathbf{X}^* \triangleq \arg\!\max_{x} \,P (\mathcal{X} | \mathcal{Z}) =  \arg\!\min_{x} \, -\!\log P(\mathcal{X}, \mathcal{Z})
\label{eq: factor_map}
\end{equation}

%%%%%%%%%%%%%%%%%%%%%%%%%%%%%%%%%%%%%%%%%%%%%%%%%%%%%%%%%%%%%%%%%%%%%%%%%%%%%%%%
% Methods
\section{Proposed Calibration Approach}
\label{sec:approach}

\noindent We report two methods based on the novel MAgnetometer and GYroscope Calibration (\magyc) approach to estimate the complete calibration of a three-axis magnetometer, i.e., hard-iron and soft-iron, and a three-axis gyroscope using magnetometer and angular rate measurements in the instrument frame, i.e., the attitude of the instrument is not required.

%%%%%%%%%%%%%%%%%%%%%%%%%%%%%%%%%%%%%%%%%%%%%%%%%%%%%%%%%%%%%%%%%%%%%%%%%%%%%%%%
% Sensor Error Model
%
\subsection{Sensor Error Model}
\label{sec:approach.sen_model}

\noindent As detailed in Section \RomanNumeralCaps{1}, magnetometers can exhibit biases referred to as hard-iron and soft-iron during operational conditions, leading to potential inaccuracies in their measurements. These biases are assumed to remain relatively constant or change slowly over time, allowing them to be treated as constants. The magnetometer model is given by

\begin{equation}
\mathbf{m_m}(t) = \mathbf{A}(\mathbf{m_t}(t) + \mathbf{m_b}),
\label{eq: magnetometer_model}
\end{equation}

\noindent where $\mathbf{m_m}(t)  \in \mathbb{R}^3$ is the noise-free magnetic field measurement in the sensor's frame, $\mathbf{m_t}(t) \in \mathbb{R}^3$ is the noise-free magnetic field real value in the sensor's frame, $\mathbf{A} \in \mathbb{R}^{3 \times 3}$ is the soft-iron, represented by a constant fully populated PDS matrix, and $\mathbf{m_b} \in \mathbb{R}^3$ is a constant pseudo-hard-iron, that once scaled by $\mathbf{A}$ will give us the magnetometer's hard-iron.

In contrast, gyroscopes are affected by the constant sensor bias and can be represented as

\begin{equation}
\mathbf{w_m}(t) = \mathbf{w_t}(t) + \mathbf{w_b},
 \label{eq: gyroscope_model}
 \end{equation}

\noindent where $\mathbf{w_m}(t)  \in \mathbb{R}^3$ is the noise-free gyroscope measurement in the sensor's frame, $\mathbf{w_t}(t) \in \mathbb{R}^3$ is the noise-free gyroscope real value in the sensor's frame, and $\mathbf{w_b} \in \mathbb{R}^3$ is the constant gyroscope bias.

%%%%%%%%%%%%%%%%%%%%%%%%%%%%%%%%%%%%%%%%%%%%%%%%%%%%%%%%%%%%%%%%%%%%%%%%%%%%%%%%
% System Model
%
\subsection{System Model}
\label{sec:approach.sys_model}

\noindent A magnetometer measures, in instrument coordinates, the Earth’s local magnetic field, which is considered to be \textit{locally} constant and fixed with respect to the inertial world frame of reference. From (\ref{eq: magnetometer_model}), we can clear the true magnetic field ($\mathbf{m_t}(t)$) and convert it to world coordinates given a rotation matrix $\mathbf{R}$. Then, we can differentiate the equation with respect to time, removing the local magnetic field from the system's model, which yields to

\begin{equation}
\dot{\mathbf{R}}(\mathbf{A}^{-1}\mathbf{m_m}(t) - \mathbf{m_b}) + \mathbf{R}\mathbf{A}^{-1}\mathbf{\dot{m}_m}(t) = 0.
\end{equation}

Using the standard equation $\dot{\mathbf{R}}(t) = \mathbf{R}(t)\left[\mathbf{w}(t)\right]_\times$ \cite{Lynch2017}, and incorporating the gyroscope bias, we derive a more comprehensive \ac{ahrs} calibration that considers both magnetometer and gyroscope calibrations. This yields a nonlinear system model independent of the instrument's attitude $\mathbf{R}(t)$,

\begin{equation}
\left[\mathbf{w_m}(t) - \mathbf{w_b}\right]_\times (\mathbf{A}^{-1}\mathbf{m_m}(t) - \mathbf{m_b}) + \mathbf{A}^{-1}\mathbf{\dot{m}_m}(t) = 0.
\label{eq: non_linear_model_gyro_bias}
\end{equation}

If we define $\mathbf{C}$ as the inverse of the soft-iron, i.e., $\mathbf{C} = \mathbf{A}^{-1}$, then $\mathbf{C}$ will also be a symmetric matrix. Moreover, if $\mathbf{A}$ is a PDS matrix, all the eigenvalues of $\mathbf{C}$ will also be positive, as they are the reciprocals of the eigenvalues of $\mathbf{A}$; therefore, $\mathbf{C}$ is also a PDS matrix. This allow us to represent $\mathbf{C}$ using the Cholesky decomposition $\mathbf{C} = \mathbf{L}\mathbf{L}^T$, where

\begin{equation}
    C = \mathbf{L}\mathbf{L}^T, \text{ with } \mathbf{L} = \begin{bmatrix} l_0 & 0 & 0 \\ l_1 & l_2 & 0 \\ l_3 & l_4 & l_5 \end{bmatrix}.
\end{equation}

% 
% Reviewers' Reply
\tfflagr{R1-3}
To ensure $\mathbf{C}$ remains PDS, we can parameterize the diagonal elements of $\mathbf{L}$ as exponential terms. Additionally, to constrain the scale of the soft-iron matrix, we can enforce a unitary volume for the ellipsoid defined by $\mathbf{A}$, which is equivalent to constraining $\det{(\mathbf{A})} = 1$. This implies $\det{(\mathbf{L}\mathbf{L}^T)} = 1$, leading to the constraint $\exp{(l_0)} \cdot \exp{(l_2)} \cdot \exp{(l_5)} = 1$, or equivalently: $\mathbf{L}_{22} = \frac{1}{\exp{(l_0)} \cdot \exp{(l_2)}}$. Thus,

\begin{equation}
    \mathbf{C} = \mathbf{L}\mathbf{L}^T, \text{ with } \mathbf{L} = \begin{bmatrix}e^{l_0} & 0 & 0 \\ l_1 & e^{l_2} & 0 \\ l_3 & l_4 & \frac{1}{e^{l_0} \cdot e^{l_2}} \end{bmatrix}.
    \label{eq: cholesky_reparametrization}
\end{equation}

% 
% Reviewers' Reply
\tfflagr{R1-5}
From (\ref{eq: non_linear_model_gyro_bias}) and the proposed parametrization \eqref{eq: cholesky_reparametrization}, we estimate the soft-iron ($\mathbf{A}$), the pseudo-hard-iron ($\mathbf{m_b}$), therefore, the hard-iron ($\mathbf{A}\mathbf{m_b}$), and the gyroscope bias ($\mathbf{w_b}$), using gyroscope ($\mathbf{w_m}(t)$) and magnetometer ($\mathbf{m_m}(t)$) measurements. Notably, to streamline computation during optimization, we optimize for the inverse of the soft-iron matrix rather than recalculating the inverse at each step. The soft-iron matrix itself is then obtained post-optimization. This approach leverages pose graph-based optimization to solve for the calibration parameters efficiently.

%%%%%%%%%%%%%%%%%%%%%%%%%%%%%%%%%%%%%%%%%%%%%%%%%%%%%%%%%%%%%%%%%%%%%%%%%%%%%%%%
% Approach: Factor Graph
%
\subsection{Factor Graph Approach}
\label{sec:approach.fg}

\begin{figure}[t!]
\centering
\begin{tikzpicture}[node distance=1.5cm, scale=0.8]
\node (center) [draw, circle, at={(0,0)}, fill=gray!30]{$\begin{bmatrix} \mathbf{l} \\ \mathbf{m_b} \\ \mathbf{w_b} \end{bmatrix}$};
\node (node0) [draw, circle, at=({0:2.5}), minimum size=0.1] {$L_{R0}$};
\node (node1) [draw, circle, at=({-30:2.5}), minimum size=0.1cm] {$L_{R1}$};
\node (node2) [draw, circle, at=({-60:2.5}), minimum size=0.1cm] {$L_{R2}$};
\node (node3) [draw, circle, at=({-90:2.5}), minimum size=0.1cm] {$L_{R3}$};
\node (node4) [draw, circle, at=({60:2.5}), minimum size=0.1cm] {$L_{Rn}$};
\node (node5) [draw, circle, at=({195:2.5}), minimum size=0.1cm] {$L_{Ri}$};
\node (invisible0)[draw=none, at=({255:2.5})]{} ;
\node (invisible1)[draw=none, at=({210:2.5})]{} ;
\node (invisible2)[draw=none, at=({180:2.5})]{} ;
\node (invisible3)[draw=none, at=({75:2.5})]{} ;

\draw [<-, line width=0.5mm] (center) -- (node0);
\draw [<-, line width=0.5mm] (center) -- (node1);
\draw [<-, line width=0.5mm] (center) -- (node2);
\draw [<-, line width=0.5mm] (center) -- (node3);
\draw [<-, line width=0.5mm] (center) -- (node4);
\draw [<-, line width=0.5mm] (center) -- (node5);
\draw [dashed, line width=0.5mm]  (invisible0) arc (255:210:2.5cm) -- (invisible1);
\draw [dashed, line width=0.5mm]  (invisible2) arc (180:75:2.5cm) -- (invisible3);
\end{tikzpicture}
\caption{A factor graph representation, depicting the residual factor, $L_{Ri}$, and the state, $x$.}
\label{fig: factor_graph_plot}
\end{figure}
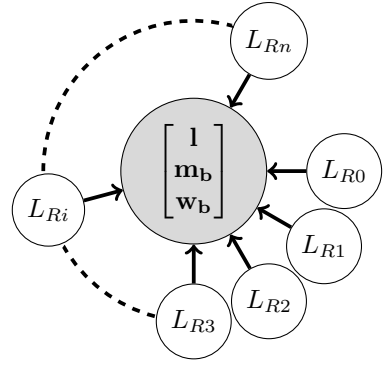

\noindent To solve for \eqref{eq: non_linear_model_gyro_bias} based on \eqref{eq: cholesky_reparametrization}, we model the system as a factor graph, we will use a single node to represent the state $\mathbf{x} = [\mathbf{l} \; \mathbf{m_b} \; \mathbf{w_b}]^T$, where, $\mathbf{l} = ( l_0 \; l_1 \; l_2 \; l_3 \; l_4)^T$ denotes the vector for the Cholesky parametrization of $\mathbf{C}$ as defined in \eqref{eq: cholesky_reparametrization}. The residual from the non-linear model (\ref{eq: non_linear_model_gyro_bias}) is represented as unary factors, as depicted in Fig. \ref{fig: factor_graph_plot}.

The decision to employ a single node instead of multiple, one for each magnetometer-gyroscope measurement set (i.e., $m$ nodes), is based on the assumption of constant calibration parameters. If multiple nodes were used, a binary factor would be necessary between each state, totaling $m - 1$ constraints to maintain the constantness of the state vector; however, this may not always be feasible due to the probabilistic nature of the state vector in a factor graph. Modeling the system as a single node ensures one set of calibration values and spares the use of the $m - 1$ constraints.

As per (\ref{eq: non_linear_model_gyro_bias}), a unary factor can be defined for the residual of each magnetometer-gyroscope measurement pair. The equation for this factor is shown in (\ref{eq: unary_factor_residual}), where $h_{r}(x)$ represents the residual model (\ref{eq: non_linear_model_gyro_bias}).

\begin{equation}
L_R(\mathbf{x};(\mathbf{z}_{mag}, \mathbf{z}_{gyro})) = \exp \left\{ -\frac{1}{2} ||h_{r}(\mathbf{x})||_\mathbf{\Sigma}^2 \right\}
\label{eq: unary_factor_residual}
\end{equation}

To apply the unary factor in \eqref{eq: unary_factor_residual}, we compute the algebraic Jacobian of the residual model. This is achieved using the Kronecker product and the vec-operator, as outlined in \eqref{eq: jacobian_residual_factor}. For brevity, we do not expand these equations further in this publication. In \eqref{eq: jacobian_residual_factor}, $i$ denotes the $i-th$ sample. Note that the differentiation of the magnetic field is not directly available and must be computed numerically.

\begin{subequations}\label{eq: jacobian_residual_factor}
\begin{align}
\frac{h_{ri}(\mathbf{x})}{\mathbf{l}} &=  (\mathbf{m_i}^T \otimes [\mathbf{w_i} - \mathbf{w_b}] + \mathbf{\dot{m}_i}^T \otimes \mathbb{I}_ 3) \frac{\partial vec(\mathbf{C})}{\partial \mathbf{l}} \\
\frac{h_{ri}(\mathbf{x})}{\mathbf{m_b}} &=  [\mathbf{w_b} - \mathbf{w_i}] \\
\frac{h_{ri}(\mathbf{x})}{\mathbf{w_b}} &= -\frac{\partial \left(\mathbf{m_i} \otimes [\mathbf{w_b}]\right)}{\partial \mathbf{w_b}} \cdot vec(\mathbf{C}) +  \frac{\partial [\mathbf{w_b}]}{\partial \mathbf{w_b}} \cdot \mathbf{m_b} 
\end{align}
\end{subequations}

%%%%%%%%%%%%%%%%%%%%%%%%%%%%%%%%%%%%%%%%%%%%%%%%%%%%%%%%%%%%%%%%%%%%%%%%%%%%%%%%
% Diagram with single node factor graph
% Description: Diagram with the single node in the middle and the factors as inputs
%

It is worth noting that the same factor \eqref{eq: unary_factor_residual} can be added to different value nodes in the graph, allowing biases to vary over time and facilitating integration with \ac{slam} graphs. However, such integration is beyond the scope of this paper.

%%%%%%%%%%%%%%%%%%%%%%%%%%%%%%%%%%%%%%%%%%%%%%%%%%%%%%%%%%%%%%%%%%%%%%%%%%%%%%%%
% Algorithm: Factor Graph
% Description: Simplified version of the factor graph algorithm
%
\begin{algorithm}[b!]
\caption{\magyc Factor Graph}\label{alg: magyc_fg}
\begin{algorithmic}
\STATE Step 1: Initialize the non-linear factor graph, the noise model
\STATE \hspace{1.0cm} for \eqref{eq: unary_factor_residual}, the average window $\theta$, and $i=0$.
\STATE Step 2: Initialize $\mathbf{l}^* = ( 0.0 \; 0.0 \; 0.0 \; 0.0 \; 0.0 )^T$, $\mathbf{m_b}^* = \vec{0}$
\STATE \hspace{1.05cm} $\mathbf{w_b}^* = \vec{0}$.
\STATE Step 3: \textbf{for each} measurement do:
\STATE \hspace{1.5cm} Accumulate measurements $\mathbf{m_i}, \dot{\mathbf{m_i}}, \mathbf{w_i}$.
\STATE \hspace{1.5cm} \textbf{if} $i \% \theta = 0$ \textbf{do}:
\STATE \hspace{2.0cm} Compute median for accumulated $\mathbf{m}, \dot{\mathbf{m}}, \mathbf{w}$.
\STATE \hspace{2.0cm} Compute residual factor error from \eqref{eq: unary_factor_residual}.
\STATE \hspace{2.0cm} Compute residual factor jacobian from \eqref{eq: jacobian_residual_factor}.
\STATE \hspace{2.0cm} Add factors to the graph.
\STATE \hspace{2.0cm} \textbf{if} iterative mode \textbf{do}:
\STATE \hspace{2.5cm} Optimize graph.
\STATE \hspace{2.0cm} Reset accumulation.
\STATE \hspace{2.0cm} i $\gets$ i + 1
\STATE Step 4: \textbf{if} batch mode \textbf{do}:
\STATE \hspace{1.5cm} Optimize graph.
\STATE Step 5: Use $\mathbf{l}^*, \mathbf{m_b}^*$, $\mathbf{w_b}^*$ to compute $\mathbf{A}, \mathbf{Am_b}, \mathbf{w_b}$.
\end{algorithmic}
\end{algorithm}

In the factor graph construction, given a set of $m$ pairs of magnetometer-gyroscope measurements, the factor graph should be composed of a single node for the state vector and $m$ factors from the system model residual \eqref{eq: unary_factor_residual}. Nevertheless, to reduce the computational load of the graph, we employ an averaging window of size $\theta$, which reduces the unary factors from $m$ to $n$, with $n = m/\theta$. By averaging $\theta$ measurements before incorporating them into the graph, we can decrease the number of factors and effectively smooth the raw measurements and the movement while filtering out high-frequency noise in the signal, preserving the underlying trend. In this paper, to ensure real-time operation during long-duration tasks, the averaging window's length is set to the sensor's frequency, allowing the factor graph to manage the same period, regardless of the sensor's frequency. As an alternative to this approach, marginalization could be used to drop measurement pairs that are not significant for the calibration algorithm. However, this implementation is beyond the scope of this work.

The factor graph's construction can follow two methods based on implementation scenarios. In the first method, calibration is performed as a post-processing step, where all $m$ measurements are initially collected, and then the optimization process incorporates all $n$ factors. The second approach involves constructing the graph incrementally, with factors added as they become available, and optimization occurs after a specified number of factors have been added.

As mentioned earlier in Section \RomanNumeralCaps{2}.E, constructing the graph requires solving a nonlinear least squares problem with sparse properties. To address this challenge, Rosen et al. \cite{Rosen2014} proposed RISE, an incremental trust-region method designed for robust online sparse least-squares estimation. As demonstrated in \cite{Rosen2014}, compared to current state-of-the-art sequential sparse least-squares solvers, RISE offers improved robustness against nonlinearity in the objective function and numerical ill-conditioning, leveraging recent advancements in incremental optimization for fast online computation.

This method was implemented using Python 3.10, using the structure outlined in the algorithm \ref{alg: magyc_fg}.

%%%%%%%%%%%%%%%%%%%%%%%%%%%%%%%%%%%%%%%%%%%%%%%%%%%%%%%%%%%%%%%%%%%%%%%%%%%%%%%%
% Figure: Simulated dataset magnetic field plots
% Description: Magnetic field plot for WAM, MAM, LAM
%
\newlength{\subfigwidth}
\setlength{\subfigwidth}{1.16in}

\begin{figure}[!t]
\centering
\subfloat[]{\includegraphics[width=\subfigwidth]{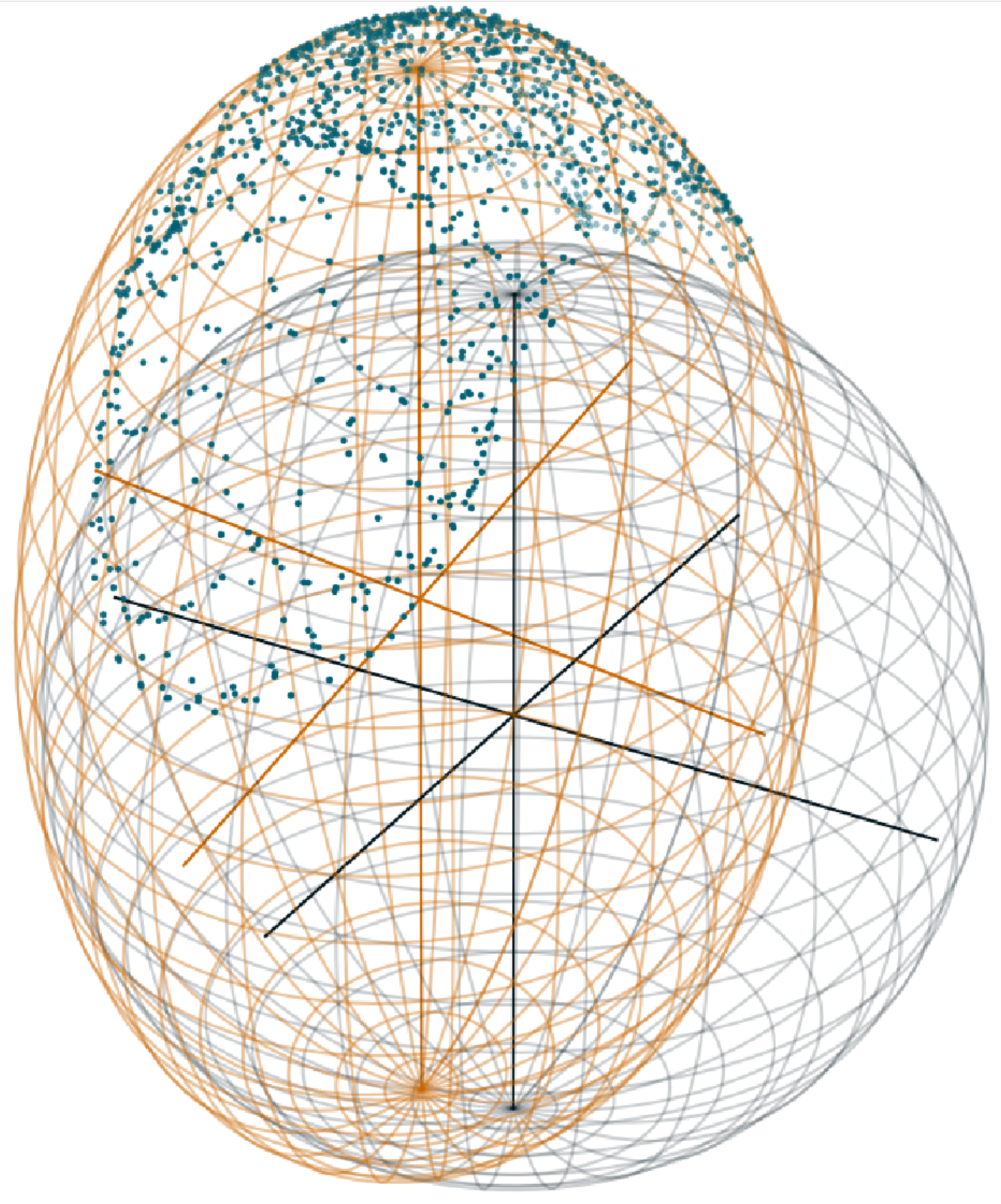}%
\label{fig: sim_data_high_plot}}
\hfil
\subfloat[]{\includegraphics[width=\subfigwidth]{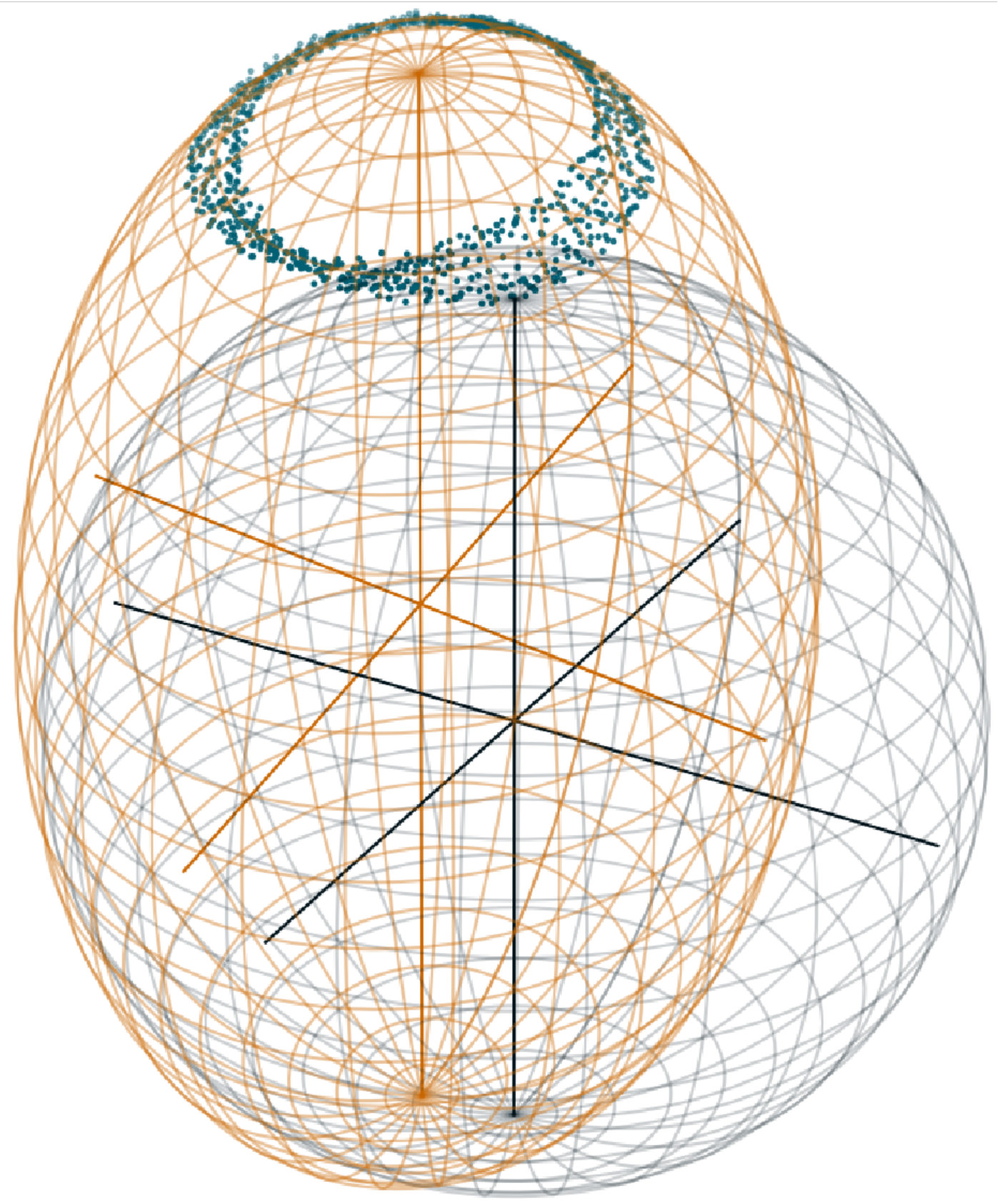}%
\label{fig: sim_data_mid_plot}}
\hfil
\subfloat[]{\includegraphics[width=\subfigwidth]{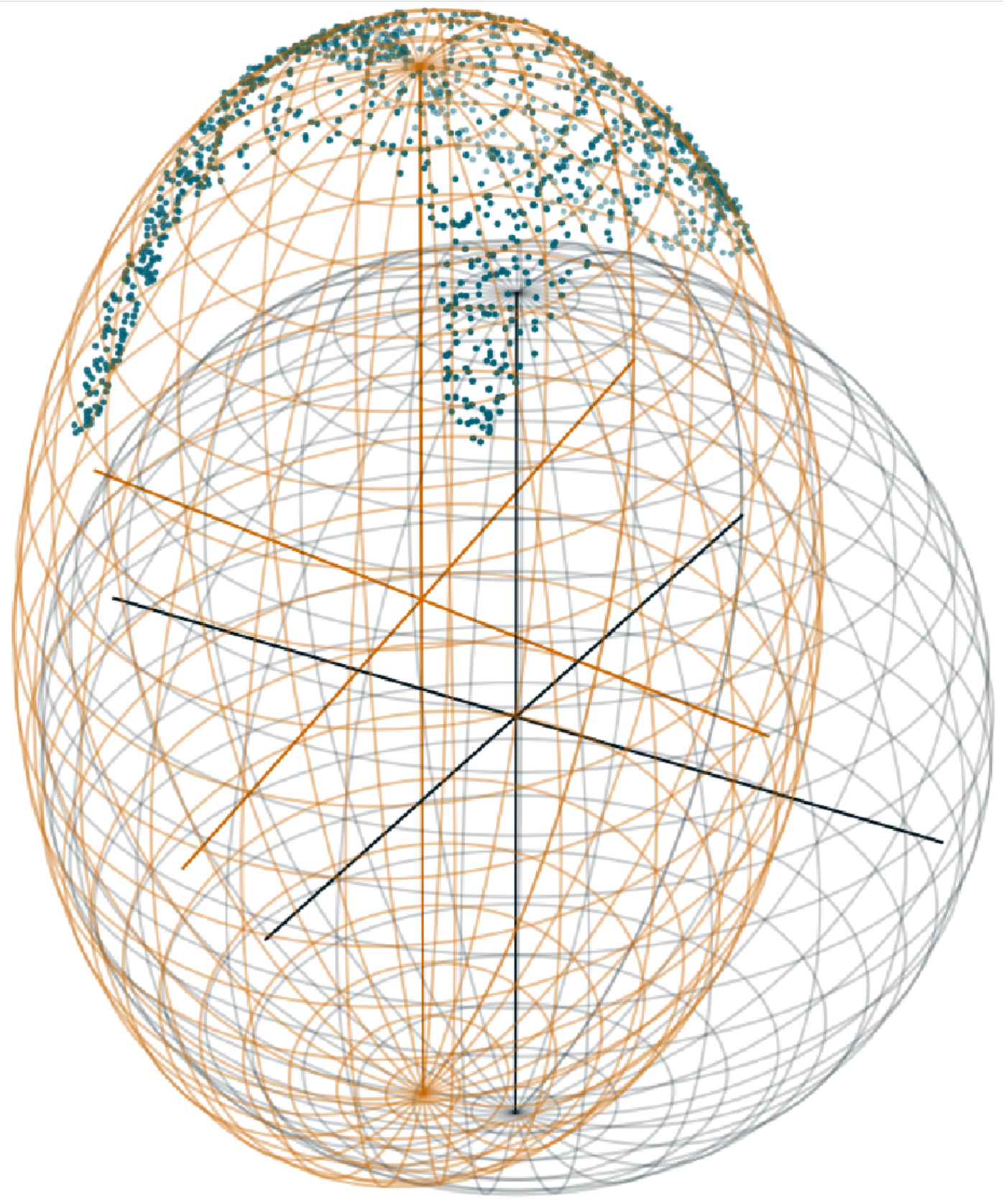}%
\label{fig: sim_data_low_plot}}
\caption{Simulated magnetometer data for three datasets: (a) \acs{wam}, (b) \acs{mam}, and (c) \acs{lam}. The 3D plots show blue dots for magnetometer data, gray spheres for the true magnetic field, and orange ellipsoids for the distorted magnetic field.}
\label{fig_sim_data_plots}
\end{figure}

%%%%%%%%%%%%%%%%%%%%%%%%%%%%%%%%%%%%%%%%%%%%%%%%%%%%%%%%%%%%%%%%%%%%%%%%%%%%%%%%
% Evaluation Methodology
\section{Evaluation Methodology}
\label{sec:evaluation}

\noindent We compared the performance of five methods for three-axis magnetometer calibration and, optionally, for three-axis gyroscope calibration. These methods can be divided into batch and real-time solutions, with the proposed methods highlighted in bold. The batch methods are as follows:

%%%%%%%%%%%%%%%%%%%%%%%%%%%%%%%%%%%%%%%%%%%%%%%%%%%%%%%%%%%%%%%%%%%%%%%
% Batch Methods
%
\begin{enumerate}
\item \textit{\textbf{\magycbfg}}: The calibration parameters are estimated using the batch mode factor graph approach described in Section \RomanNumeralCaps{3}.C, where all the factors are added to the factor graph before optimization.
\item \textit{\twostep}: The calibration parameters are estimated using the widely cited \twostep method \cite{Alonso2002a}, which is based on the implementation proposed by Dinale \cite{Dinale2013}. This method takes as input the local value of the magnetic field.
\item \textit{Ellipsoid Fit}: The calibration parameters are estimated using the widely used Ellipsoid Fit method, based on the implementation proposed by Bazhin et al. \cite{Bazhin2022}.
\end{enumerate}

\noindent The real-time methods are listed below:

%%%%%%%%%%%%%%%%%%%%%%%%%%%%%%%%%%%%%%%%%%%%%%%%%%%%%%%%%%%%%%%%%%%%%%%
% Real Time Methods
%
\begin{enumerate}
\item \textit{\textbf{\magycifg}}: The calibration parameters are estimated using the incremental mode factor graph approach described in Section \RomanNumeralCaps{3}.C, where the factors are added as they are received.
\item \textit{\magfactor}: The calibration parameters are estimated using a factor graph approach that uses the magnetic calibration factor included in the GTSAM library \cite{BORGLab2023}. Unlike a full soft-iron matrix estimation, this method only estimates a single scale factor that is uniform across all three axes, as well as the hard-iron, being considered as an extension to the method proposed by Lathrop et al. \cite{Lathrop2023}. The current attitude of the system and the local magnetic field value must be provided as inputs.
\end{enumerate}

%%%%%%%%%%%%%%%%%%%%%%%%%%%%%%%%%%%%%%%%%%%%%%%%%%%%%%%%%%%%%%%%%%%%%%%
% Figure: Simulated data matrix error
% Description: SI, HI and Wb matrix error with respect to the ground truth
%
\begin{figure*}[t!]
\centering\includegraphics[width=\linewidth]{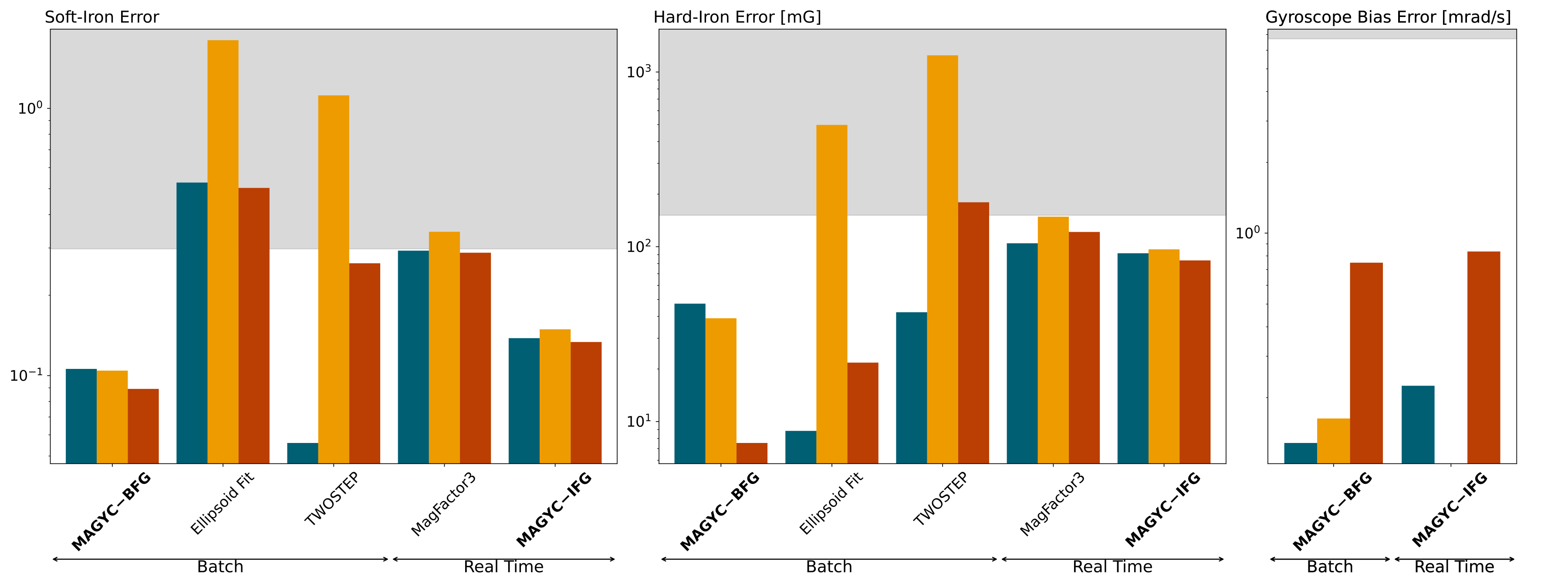}%
\caption{Performance comparison of seven calibration methods on three simulated datasets. The geodesic soft-iron error $\log{(||A^{-1/2} A^* A^{-1/2}||_F)}$, hard-iron error $|m_b - m_b^*|$, and gyroscope bias error $|w_b - w_b^*|$ are analyzed for the \acs{wam} (green), \acs{mam} (yellow), and \acs{lam} (orange) datasets. Gray-shaded zones show the raw data value.}
\label{fig: sim_matrix_error}
\end{figure*}

To compare batch and real-time methods, the calibration parameters estimated with the real-time methods were based on the average of the last \SI{20}{\percent} of the estimated parameters. It should be noted that the \twostep and \magfactor methods require knowledge of the local magnetic field magnitude, which was obtained from the \ac{wmm} provided by the \ac{noa} \cite{NOAA2023} for in-field evaluations.

For both numerical and in-field evaluations, for the factor graph-based methods, the termination criteria were set empirically, with both the relative and absolute error tolerances set to $1.0 \times 10^{-7}$. A multifrontal Cholesky factorization was used, as it has been shown to outperform the LDL and QR factorizations \cite{Dellaert2006}. The optimization was computed using the RISE method from GTSAM. The initial conditions for the state vector are the assumption of both magnetometer and gyroscope to be calibrated, i.e., \mbox{$\mathbf{m_b} = \vec{\mathbf{0}}$}, \mbox{$\mathbf{w_b} = \vec{\mathbf{0}}$} and \mbox{$\mathbf{l} = ( 0.0 \; 0.0 \; 0.0 \; 0.0 \; 0.0 )^T$}.

The factor graph was optimized each time a new pair of factors was added from an averaged sample set $i, \;\forall \; i \; \in \; \{0, \dots, n\}$, with an average window size $\theta$ set to match the sensor's frequency. The noise covariance matrices was set to $\mathbb{I}_3 \cdot 1\times10^{-6}$ for \eqref{eq: unary_factor_residual}. These values were determined through a sensitivity analysis, which found the optimal trade-off for the matrices. Due to space constraints, the details of this analysis are not included.

The code implementation of the \magyc methods, the simulation data used in Section \ref{sec:sim_results}, and example code demonstrating how to apply the calibration methods are available on the \magyc project page: \href{https://compas-docs.github.io/magyc/}{https://compas-docs.github.io/magyc/}.

%%%%%%%%%%%%%%%%%%%%%%%%%%%%%%%%%%%%%%%%%%%%%%%%%%%%%%%%%%%%%%%%%%%%%%%%%%%%%%%%
% Simulation Results
\section{Numerical Simulation Evaluation}
\label{sec:sim_results}

%%%%%%%%%%%%%%%%%%%%%%%%%%%%%%%%%%%%%%%%%%%%%%%%%%%%%%%%%%%%%%%%%%%%%%%%%%%%%%%%
% Simulation Characteristics
%
\noindent A Monte Carlo numerical simulation was conducted to replicate 6,000 measurements from a \ac{mems} \ac{ahrs} during the sinusoidal motions of a vehicle. This simulation was designed to emulate a magnetometer calibration platform for an articulated vehicle or one with pitching capabilities in the \ac{wam} dataset (Fig. \ref{fig: sim_data_high_plot}), a magnetometer calibration for a roll-and-pitch stable vehicle in the \ac{mam} dataset (Fig. \ref{fig: sim_data_mid_plot}), and a survey for a vehicle with the same capabilities as described in \ac{wam} for the \ac{lam} dataset (Fig. \ref{fig: sim_data_low_plot}). To generate these datasets, the motion was modeled as

\begin{equation}
    \{rph\}_i = A_i \cdot \sin{\left(\frac{\omega_i}{A_i} \cdot t + \phi_i\right)},
\end{equation}

\noindent where $A_i$ represents the angular motion amplitude for the different datasets, as specified in Table \ref{tab: sim_angular_range}. The angular rates of the vehicle ($\omega_i$) are randomly initialized based on a uniform distribution within the following ranges: 0.05 to \SI{0.08}{\radian\per\second} for roll, 0.1 to \SI{0.3}{\radian\per\second} for pitch, and 0.2 to \SI{0.4}{\radian\per\second} for heading. $\phi_i$ represents a random phase shift modeled as a uniform distribution between $-\pi$ and $\pi$, to add randomness to the initial attitude.

\begin{table}[b!]
    \renewcommand{\arraystretch}{1.3}
    \caption{Simulation Dataset Roll, Pitch and Heading Motion Amplitude}\label{table: simulation_results}
    \centering
    \begin{tabular}{c@{\hspace{10mm}} c@{\hspace{10mm}} c@{\hspace{10mm}} c}
        % & \multicolumn{3}{c}{\textbf{Angular Motion Amplitude}} \\
        \textbf{Motion Level} & \textbf{Roll} (\SI{}{\degree}) & \textbf{Pitch} (\SI{}{\degree}) & \textbf{Heading} (\SI{}{\degree}) \\
        \Xhline{1.5pt}
        WAM & $\pm 5$ & $\pm 45$ & $\pm 360$ \\
        MAM & $\pm 5$ & $\pm 5$ & $\pm 360$ \\
        LAM & $\pm 5$ & $\pm 45$ & $\pm 90$ \\
        \Xhline{1.5pt}
    \end{tabular}
    \label{tab: sim_angular_range}
\end{table}

Each experiment lasted \SI{600}{\second}, with simulated data generated at a \SI{10}{\hertz} rate and magnetometer measurements ($\sigma_{mag} = \SI{10}{\milligauss}$) and angular rate sensor ($\sigma_{gyro} = \SI{10}{\milli\radian\per\second}$) corrupted by Gaussian noise.

The true magnetic field vector is $\mathbf{m_0} = [227,\, 52, \,412]^T \SI{}{\milligauss}$, the soft-iron upper triangular terms are given by $\mathbf{a} = [1.10,\, 0.10,\, 0.04,\, 0.88,\, 0.02,\, 1.22]^T$, the hard-iron bias is $\mathbf{m_b} = [20,\, 120,\, 90]^T \SI{}{\milligauss}$, and the gyroscope bias is $\mathbf{w_b} = [4,\, -5,\, 2]^T \SI{}{\milli\radian\per\second}$. The magnetic field used in the \twostep and \magfactor methods was randomly scaled using a Gaussian distribution, where $\mu = 1.0$ and $\sigma = 0.05$, relative to the simulated data.

As evaluation metrics, given the availability of ground truth values, we compute the matrix error for the soft-iron matrix using the geodesic distance \eqref{eq: geodesic_distance} and use the Euclidean distance for both the hard-iron and gyroscope biases, as summarized in Fig. \ref{fig: sim_matrix_error}. To assess the calibration performance, we also calculate the standard deviation of the magnetic field and the RMSE for the heading estimation, with results summarized in Table \ref{table: simulation_results}.

%%%%%%%%%%%%%%%%%%%%%%%%%%%%%%%%%%%%%%%%%%%%%%%%%%%%%%%%%%%%%%%%%%%%%%%%%%%%%%%%
% Table: Simulation results table
% Description: Mean heading RMSE and Magnetic field std for WAM, MAM and LAM
%
\begin{table*}[!t]
\renewcommand{\arraystretch}{1.3}
\caption{Mean heading RMSE and magnetic field standard deviation metrics for three batch and two real-time calibration methods across three simulated datasets, evaluated over 100 validation simulations. The best two results in each column are \textbf{bolded}. The percentage superscript indicates the failure rate to converge to a solution in a subset of simulations}\label{table: simulation_results}
\centering
\begin{tabular}{p{0.05cm} p{0.1cm} l c@{\hspace{5mm}}c c@{\hspace{5mm}}c c@{\hspace{5mm}}c c@{\hspace{5mm}}c}
\multicolumn{2}{c}{\multirow{4}[4]{*}{}} &
\multicolumn{1}{c}{\multirow{4}[4]{*}{}} &
\multicolumn{2}{c}{\textbf{WAM for Calibration}} &
\multicolumn{2}{c}{\textbf{MAM for Calibration}} &
\multicolumn{2}{c}{\textbf{LAM for Calibration}}  \\
\Xhline{1.5pt}
& & & Mean Heading & Magnetic Field & Mean Heading & Magnetic Field & Mean Heading & Magnetic Field \\
& & & RMSE (\unit{\degree}) & Std (\unit{\milligauss}) & RMSE (\unit{\degree}) & Std (\unit{\milligauss}) & RMSE (\unit{\degree}) & Std (\unit{\milligauss}) \\
\hline
\multicolumn{2}{c}{}  & Raw & 39.067 & 52.426 & 39.067 & 52.426 & 39.067 & 52.426 \\
\hline
\multicolumn{2}{c}{\parbox[t]{2mm}{\multirow{3}{*}{\rotatebox[origin=c]{90}{BATCH}}}} & \magycbfg & \textbf{13.160} & \hspace{1mm}\textbf{9.668} & \textbf{13.176} & \hspace{1.5mm}\textbf{9.875} & \textbf{13.125} & \hspace{1mm}\textbf{9.354} \\

\multicolumn{2}{c}{}                                      & \twostep & \textbf{13.200} & \hspace{1mm}\textbf{8.637} & \hspace{3mm}697.762\textsuperscript{78\%} & \hspace{4mm}71.464\textsuperscript{78\%} & \hspace{4mm}14.971\textsuperscript{17\%} & \hspace{5mm}\textbf{8.612}\textsuperscript{17\%} \\

\multicolumn{2}{c}{}                                      & Ellipsoid Fit & 20.640 & 47.238 & \hspace{2mm}593.282\textsuperscript{7\%} & \hspace{2mm}342.394\textsuperscript{7\%} & 19.636 & 44.034 \\
\hline
\parbox[t]{2mm}{\multirow{2}{*}{\rotatebox[origin=c]{90}{REAL}}} & \parbox[t]{2mm}{\multirow{2}{*}{\rotatebox[origin=c]{90}{TIME}}} & \magycifg & 13.306 & 10.472 & \textbf{13.420} & \textbf{10.844} & \textbf{13.148} & 10.955 \\
\multicolumn{2}{c}{}                                             & \magfactor & 27.811 & 33.979 & 26.111 & 41.069 & 31.097 & 36.937 \\
\Xhline{1.5pt}
\end{tabular}
\end{table*}

%%%%%%%%%%%%%%%%%%%%%%%%%%%%%%%%%%%%%%%%%%%%%%%%%%%%%%%%%%%%%%%%%%%%%%%%%%%%%%%%
% Analysis results
%
The calibration methods presented in Section \RomanNumeralCaps{4} were calibrated on the three datasets described previously and subsequently evaluated on the \ac{wam} dataset. Results in Table \ref{table: simulation_results} and Fig. \ref{fig: sim_matrix_error} show that the proposed \magyc methods consistently outperformed the benchmark methods. Specifically, in the \ac{wam} dataset calibration, while the \twostep method demonstrated superior performance in soft-iron estimation, both \magycbfg and \magycifg exhibited comparable performance in terms of mean heading RMSE and magnetic field standard deviation.

In the \ac{mam} dataset, \twostep failed to converge in \SI{78}{\percent} of cases, and along with the Ellipsoid Fit method, performance was significantly degraded due to limited pitch motion, highlighting its limitation to wide-range movement scenarios, which may not always be feasible for full-scale vehicles. Meanwhile, \magfactor displayed greater robustness but did not compute non-orthogonality or scale factors along all three axes. In contrast, the proposed \magyc methods consistently demonstrated better overall performance and converged to a solution in all simulations.

In the \ac{lam} dataset, despite the heading being constrained to $\pm 90$\SI{}{\degree}, the pitch motion allowed \twostep to converge in \SI{83}{\percent} of cases and improved Ellipsoid Fit results relative to \ac{mam}. However, both \magyc methods still outperformed or matched the benchmark methods while delivering more reliable performance.

These results suggest that the \magyc methods are more resilient to constrained angular movements than state-of-the-art methods while outperforming or matching the performance on the datasets that are well suited for the benchmark methods due to the wide coverage of the magnetic field.

Additionally, the \magycbfg method demonstrates greater robustness compared to \magycifg, as anticipated given its utilization of the entire dataset in a post-processing step, in contrast to the incremental approach employed by \magycifg. For the real-time \magycifg method, we can analyze solution convergence based on a window size of 10 iterations with a relative tolerance of $1\times10^{-3}$ between components within the window to determine the convergence of the calibration parameters. This analysis shows that the soft-iron converges after receiving approximately \SI{14}{\percent} of the samples, the hard-iron after \SI{10}{\percent}, and the gyroscope bias after \SI{6}{\percent}, demonstrating rapid convergence to optimal values.

Regarding processing time, the \magycifg method has a computation time of approximately \SI{185}{\milli\second} per averaging window, demonstrating its feasibility for real-time operation by remaining within \SI{93}{\percent} of the time between optimizations, which is equivalent to \SI{1}{\second} when the proposed window size matches the sensor frequency. In contrast, the \magycbfg method exhibits a slight increase in processing time as movement range decreases, rising from approximately \SI{34.7}{\second} to \SI{55.7}{\second} per post-processing calibration due to additional iterations required to meet the termination criteria.

These simulation results support the proposed \magyc methods as competitive or superior to benchmark methods, underscoring their effectiveness for both post-processing and real-time navigation applications.

%%%%%%%%%%%%%%%%%%%%%%%%%%%%%%%%%%%%%%%%%%%%%%%%%%%%%%%%%%%%%%%%%%%%%%%
% Figure: Field data magnetic field
% Description: Magnetic field plot for EXP1 and EXP2
%
\begin{figure}[b!]
\centering
\subfloat[]{\includegraphics[width=1.16in]{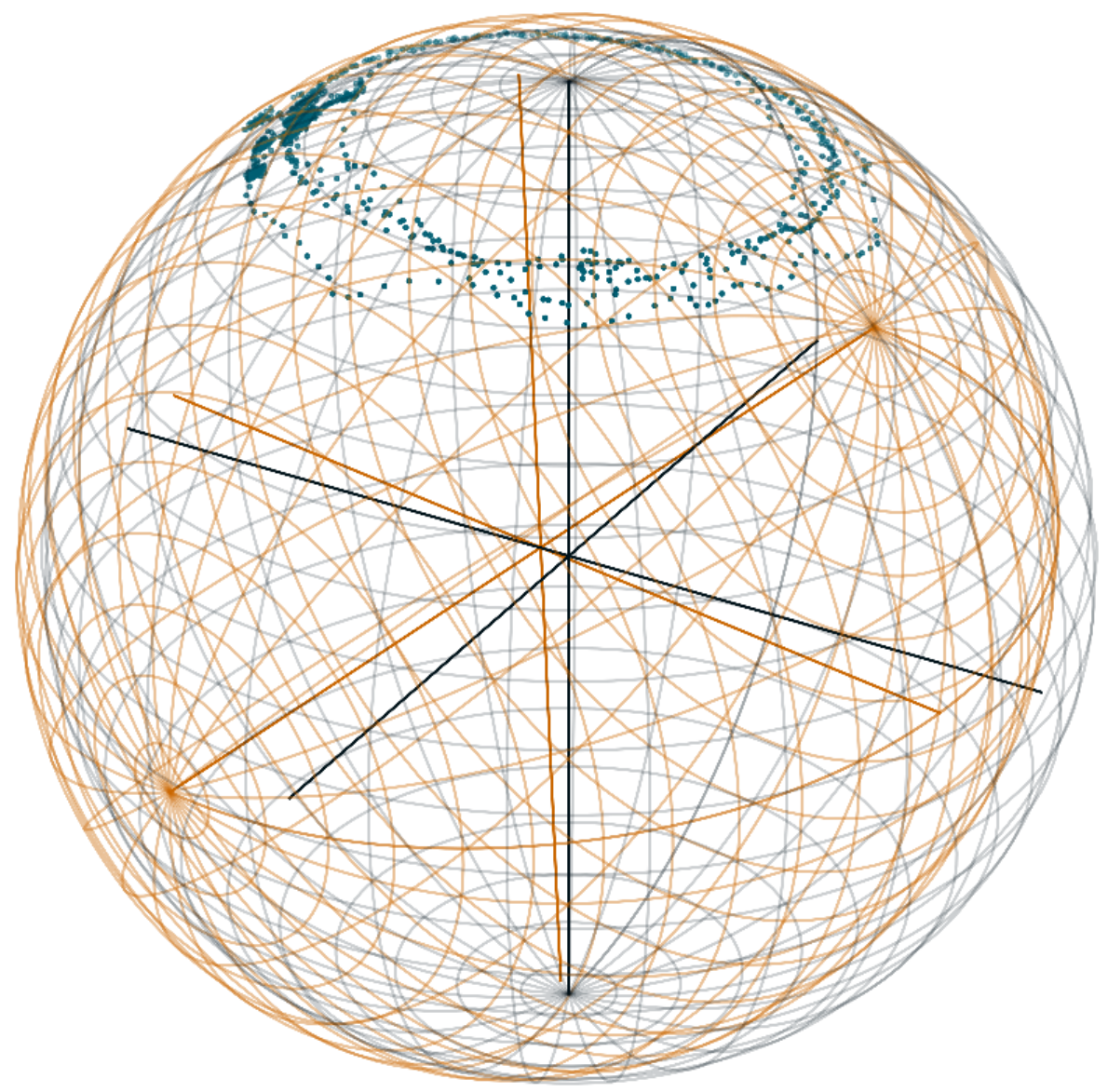}%
\label{fig: exp1_plot}}
\hfil
\subfloat[]{\includegraphics[width=1.16in]{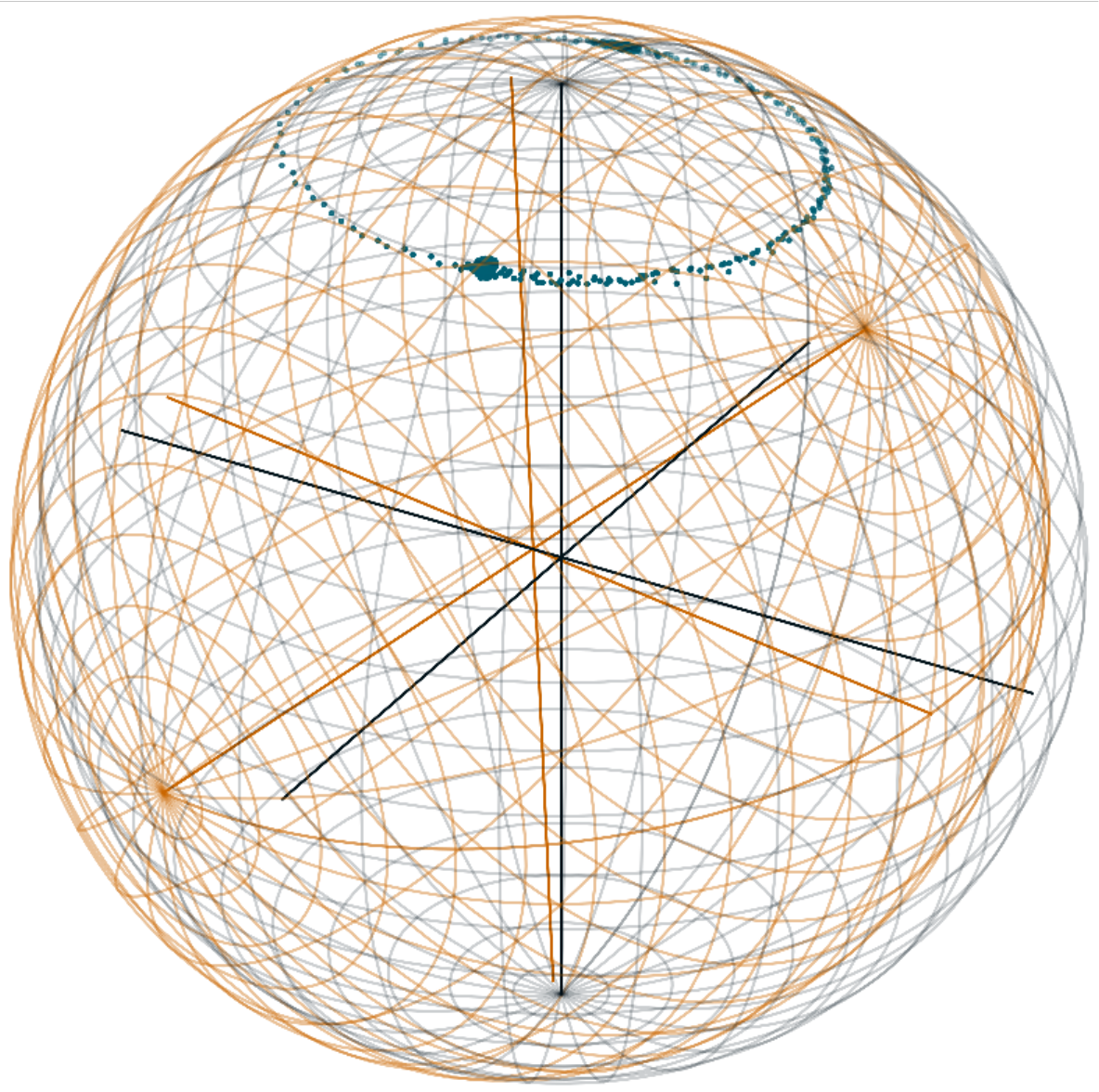}%
\label{fig: exp2_plot}}
\caption{In-field magnetic data for two datasets: (a) EXP1-14 and (b) EXP2-14. The 3D plots show blue dots for magnetometer data, gray spheres for true magnetic field based on the magnetic model magnitude, and the orange sphere representing the non-calibrated magnetic field.}
\label{fig_exp_data_plots}
\end{figure}

%%%%%%%%%%%%%%%%%%%%%%%%%%%%%%%%%%%%%%%%%%%%%%%%%%%%%%%%%%%%%%%%%%%%%%%%%%%%%%%%
% Field Results
\section{Field Experimental Evaluation}
\label{sec:field_results}

%%%%%%%%%%%%%%%%%%%%%%%%%%%%%%%%%%%%%%%%%%%%%%%%%%%%%%%%%%%%%%%%%%%%%%%%%%%%%%%%
% Figure: Field results EXP1-14, EXP1-24 and EXP2-14
% Description: Heading and xy trajectory errors
%
\begin{figure*}[t!]
\centering
\subfloat[]{\includegraphics[width=0.5\linewidth]{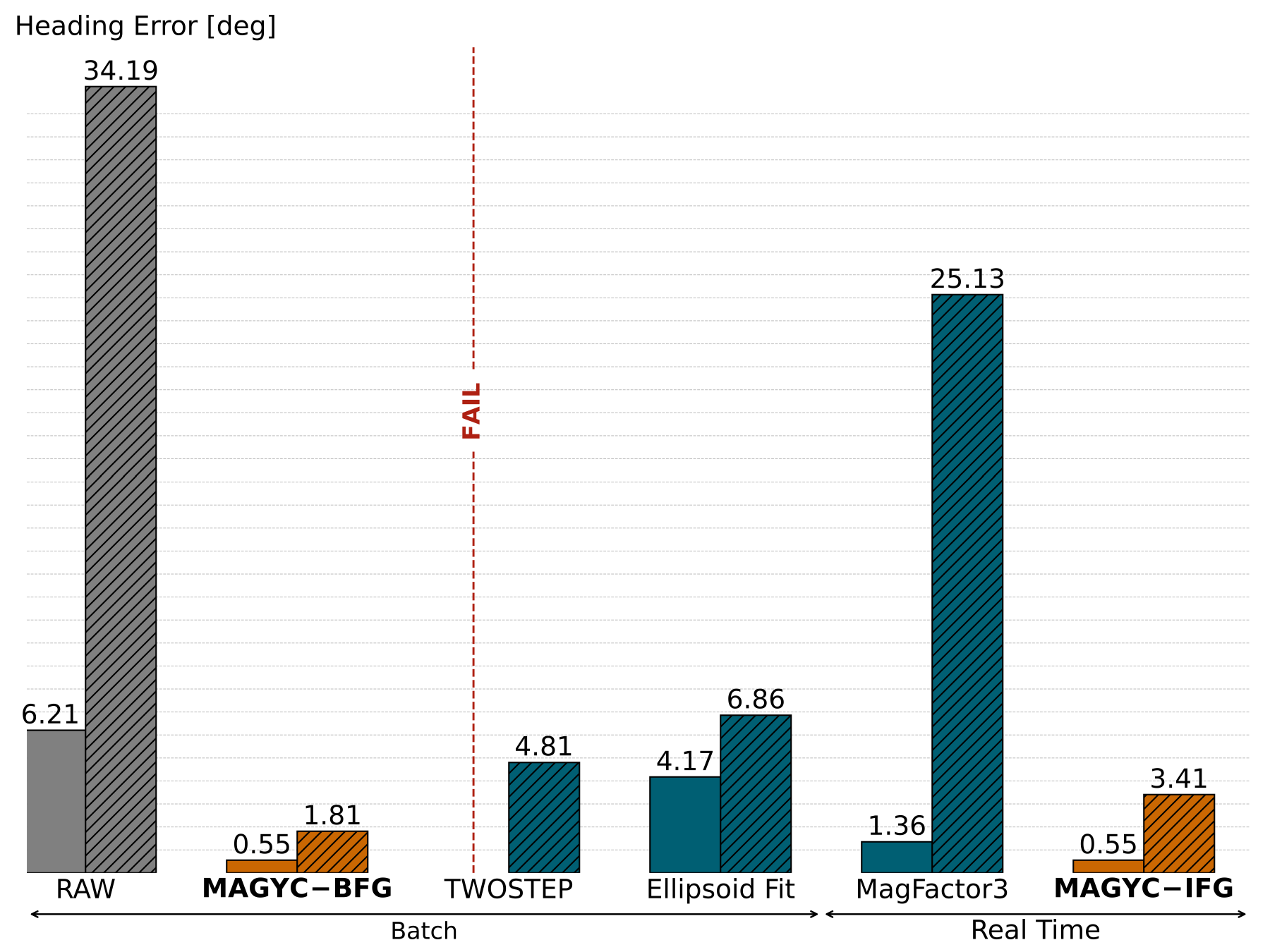}%
\label{fig: field_heading_error}}
\hfil
\subfloat[]{\includegraphics[width=0.5\linewidth]{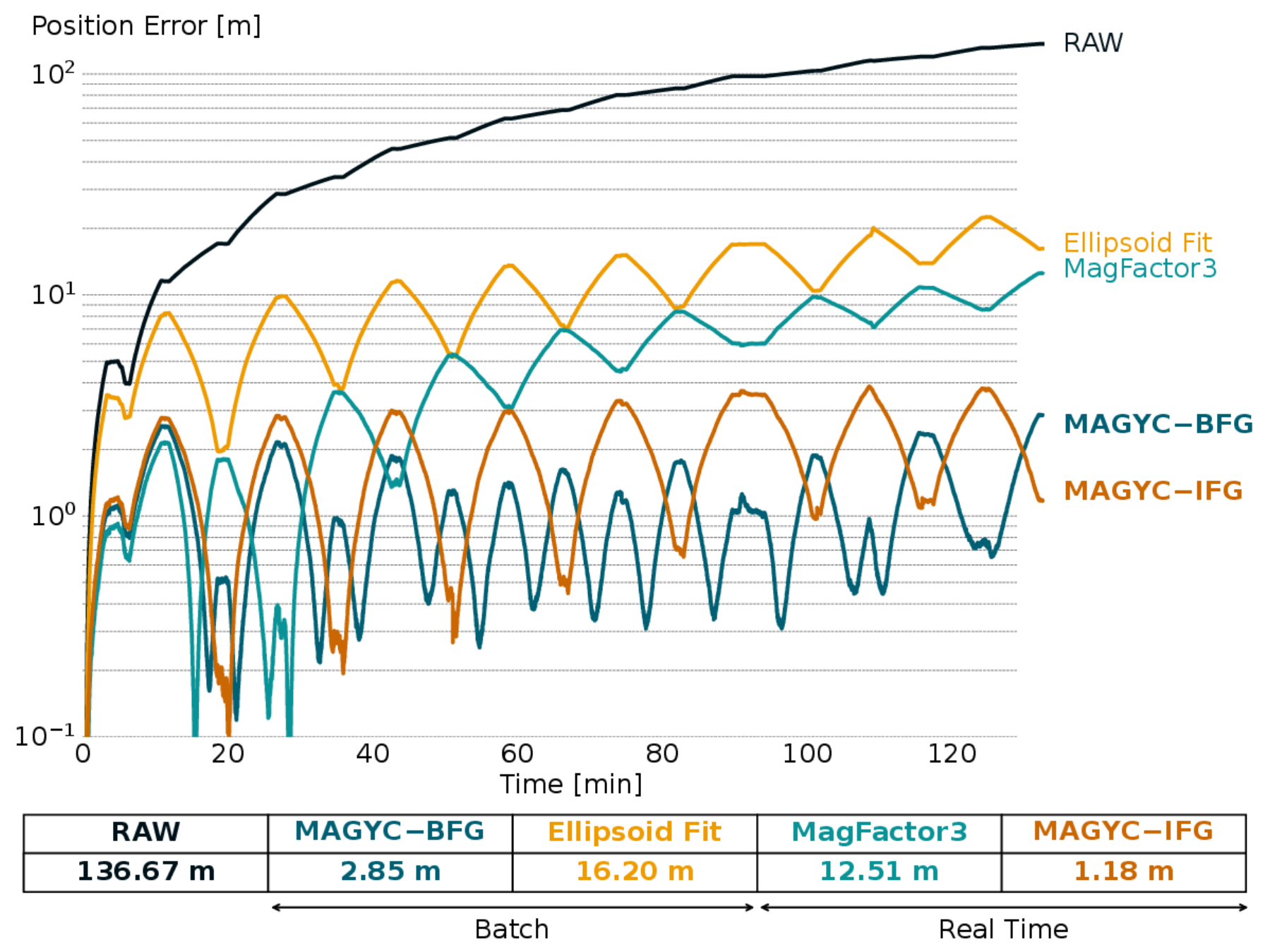}%
\label{fig: field_xy_error_exp2_14}}
\caption{Field experiment: (a) The heading error for calibration parameters estimated in experiments EXP1-14 (solid bars) and EXP1-24 (hatched bars) is evaluated in EXP2-14 and EXP2-24, respectively. Red dashed lines indicate instances where parameter estimation failed. (b) Norm of XY position error for EXP2-14, with method-wise total error, summarized in the table below.}
\label{fig: field_results}
\end{figure*}

\begin{figure}[b!]
\centering
\subfloat[]{\includegraphics[width=0.59\linewidth]{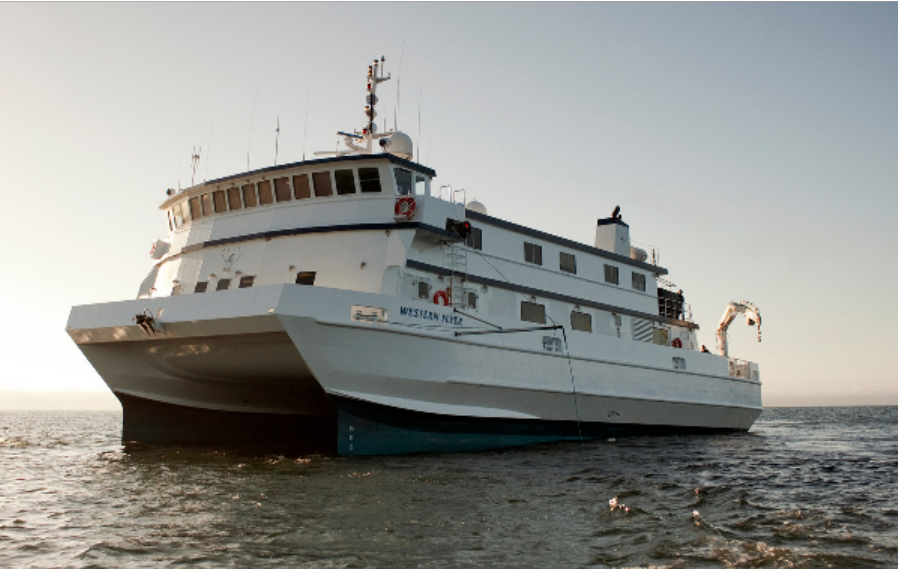}%
\label{fig:western_flyer}}
\hfill
\subfloat[]{\includegraphics[width=0.41\linewidth]{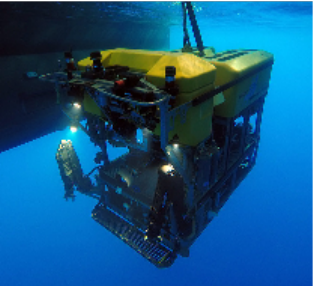}%
\label{fig:doc_ricketts}}
\caption{\acs{mbari}'s R/V Western Flyer (a) and ROV Doc Ricketts (b) deployed during the December 2014 Monterey Bay seafloor mapping expedition.}
\label{fig:mbari_equipment}
\end{figure}

%%%%%%%%%%%%%%%%%%%%%%%%%%%%%%%%%%%%%%%%%%%%%%%%%%%%%%%%%%%%%%%%%%%%%%%%%%%%%%%%
% The hardware and the experiments
%
We evaluated the in-field performance of the proposed and benchmark methods using two seafloor mapping survey dives conducted in 2014 and 2024 in Monterey Bay, where the local magnetic field had a magnitude of \SI{479}{\milligauss} \cite{NOAA2023}. The ROV Doc Ricketts was deployed from the R/V Western Flyer (Fig. \ref{fig:mbari_equipment}) in 2014, and the ROV Ventana was deployed from the R/V Rachel Carson in 2024; both ROVs and vessels are owned and operated by \ac{mbari}. Each ROV was equipped with the Low Altitude Survey System (LASS), featuring a \ac{mems} \ac{imu}, a high-end \ac{ins}, and a suite of other sensors for seafloor mapping. The main difference between the datasets is that in 2014, the LASS lacked pitch control, keeping it at a fixed pitch of \SI{0}{\degree} in 2024; however, the upgraded LASS was fitted with actuators to enable platform pitch control.
% Reviewers' Reply
\tfflagr{R2-4}
% 

%%%%%%%%%%%%%%%%%%%%%%%%%%%%%%%%%%%%%%%%%%%%%%%%%%%%%%%%%%%%%%%%%%%%%%%%%%%%%%%%
% Figure: MBARI equipment
% Description: R/V Western Flyer and ROV Doc Ricketts
%

\begin{figure}[b!]
    \centering
    \subfloat[]{\includegraphics[width=0.48\linewidth]{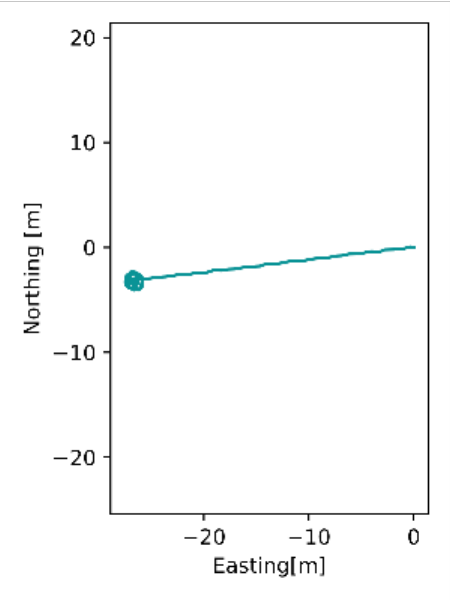}%
    \label{fig: trajectory_exp1}}
    \hfil
    \subfloat[]{\includegraphics[width=0.48\linewidth]{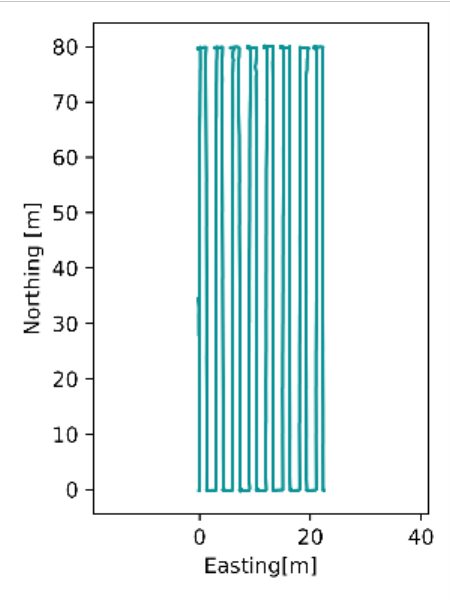}%
    \label{fig: trajectory_exp2}}
    \caption{Seafloor mapping survey field experiments trajectories using \acs{ins} data: (a) magnetometer calibration (EXP1) and (b) standard survey (EXP2).}
    \label{fig:field_trajectories}
\end{figure}

In both deployments, two field experiments were conducted during the survey. The first, labeled EXP1-14 (Figs. \ref{fig: exp1_plot} and \ref{fig:field_trajectories}) and EXP1-24, involved magnetometer calibration procedures with the platform at a fixed pitch of \SI{0}{\degree}. This experiment involved a series of \SI{360}{\degree} heading rotations, with pitch and roll oscillating around \SI{5}{\degree} due to vehicle dynamics. The second experiment, labeled EXP2-14 (Figs. \ref{fig: exp2_plot} and \ref{fig:field_trajectories}) and EXP2-24, consisting of a seafloor mapping survey with a ``mowing the lawn'' pattern. In EXP2-14, the platform was fixed, while in EXP2-24, it actively followed the seafloor.

To evaluate heading estimation performance and its impact in navigation, we used a Vectornav VN100 \acs{mems}-based \ac{imu} operating at a sampling rate of \SI{80}{\hertz}, with magnetometer noise of $\sigma_{mag} \! = \! \SI{1}{\milligauss}$ and gyroscope noise of $\sigma_{gyro} \! = \! \SI{0.5}{\milli\radian\per\second}$ \cite{vn100}. While magnetometers can be calibrated beforehand, local perturbations may induce unknown magnetometer biases once mounted on the vehicle. Thus, we utilized a Kearfott SeaDeViL high-end \ac{ins} as ground truth, operating at \SI{25}{\hertz}, equipped with a \ac{dvl} and a ring-laser gyro, offering a heading precision of \SI{0.05}{\degree} and pitch/roll precision of \SI{0.03}{\degree}, with real-time position accuracy of \SI{0.05}{\percent} of the total \ac{dt} when the \ac{dvl} continuously tracks the seafloor \cite{kearfott}.

%%%%%%%%%%%%%%%%%%%%%%%%%%%%%%%%%%%%%%%%%%%%%%%%%%%%%%%%%%%%%%%%%%%%%%%%%%%%%%%%
% Heading Error Results
%
\subsection{Heading Estimation Performance}
\label{sec:field_results.heading}

%
% Reviewers' Reply
\tfflagr{R2-5}
We interpolated the \ac{mems} \ac{imu} data to match the \ac{ins} sampling rate to estimate the vehicle's heading. For data integration, we used the Hua Complementary Filter \cite{hua2011}, a non-linear attitude estimation technique that combines accelerometer, gyroscope, and magnetometer measurements, effectively decoupling \ac{imu} readings and utilizing the local magnetic field without requiring additional magnetic deviation corrections. Based on this filter, we calculated the heading and estimated the heading error as the RMSE between the \ac{ins}' heading and the computed heading from bias-compensated magnetometer data for each evaluated method. For reference, the uncalibrated heading is labeled as “RAW” in the plots.

%
% Reviewers' Reply
\tfflagr{R2-6}
As shown in Fig. \ref{fig: field_heading_error}, the proposed \magyc methods significantly enhance heading accuracy when calibrated using EXP1-14 and EXP1-24. The initial heading error is reduced from \SI{6.21}{\degree} to \SI{0.55}{\degree} in EXP1-14 and from \SI{34.19}{\degree} to below \SI{3.41}{\degree} in EXP1-24. In EXP1-14, the \twostep method failed to converge to a solution, whereas the proposed methods outperformed both Ellipsoid Fit and \magfactor. Similarly, in EXP1-24, \magyc methods demonstrated superior performance compared to all three benchmark methods. These results highlight the robustness of the \magyc approaches across distinct applications using the same calibration strategy.

%%%%%%%%%%%%%%%%%%%%%%%%%%%%%%%%%%%%%%%%%%%%%%%%%%%%%%%%%%%%%%%%%%%%%%%%%%%%%%%%
% Figure: Mapping survey trajectories
% Description: XY trajectories for EXP1-14 and EXP2-14
%

%%%%%%%%%%%%%%%%%%%%%%%%%%%%%%%%%%%%%%%%%%%%%%%%%%%%%%%%%%%%%%%%%%%%%%%%%%%%%%%%
% Dead Reckoning Error Results
%
\subsection{Navigation Performance}
\label{sec:field_results.dr}

In underwater vehicle navigation, the most common \ac{dr} method combines \ac{dvl} velocity measurements with data from an \ac{ahrs}. Biases in \ac{ahrs} measurements can lead to heading errors, which critically impact position accuracy \cite{Troni2012}. To mitigate these issues, we applied the calibration from EXP1-14 and EXP1-24. For each method, we computed \ac{dr} using the attitude from the Hua Filter (as described in the previous section) and the velocity data from the \ac{ins}, which provided ground-truth positioning for the vehicle. We integrated these sources using a Kalman Filter, setting the process noise covariance to $k_1 = 0.021$ and the measurement noise covariance to $k_2 = 1.1$, values selected through empirical tuning. We then computed the Cartesian error with respect to the ground-truth positions, providing a comparative measure of each method's accuracy.

Fig. \ref{fig: field_xy_error_exp2_14} illustrates the norm of the XY position error for each calibration method using the EXP2-14 dataset, where the platform was fixed at an angle aligned with gravity. Without any magnetometer calibration (RAW), the position error after traveling \SI{1395}{\meter} reaches \SI{137}{\meter}, accounting for a substantial \SI{10}{\percent} error. In contrast, the proposed \magyc methods show an improvement, with position errors reduced to under \SI{2.85}{\meter}, corresponding to just \SI{0.20}{\percent} of the total \ac{dt}. Benchmark methods, including Ellipsoid Fit and \magfactor, display errors that are an order of magnitude larger than those of the \magyc methods. Furthermore, the \twostep method fails to converge due to its inherent limitations in scenarios with minimal movement, highlighting the robustness and reliability of the proposed \magyc approaches.

\begin{figure}[t!]
\centering
\includegraphics[width=\linewidth]{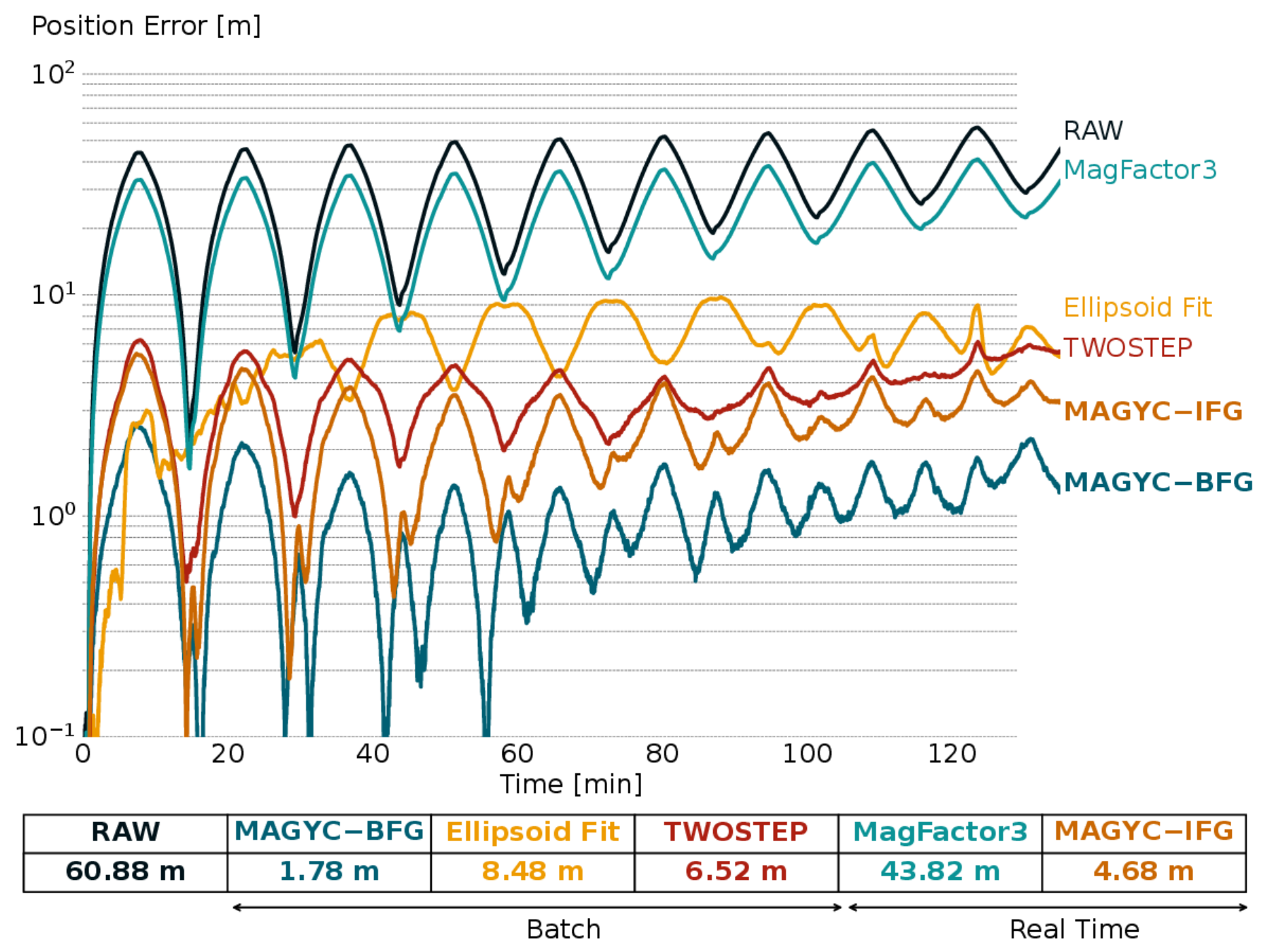}%
\label{fig: field_xy_error}
\caption{Field experiment: Norm of XY position error for EXP2-24, with method-wise total error, summarized in the table below.}
\label{fig: field_results_exp2_24}
\end{figure}

When analyzing the trajectory results produced by \ac{dr} for each method, navigation becomes infeasible without magnetometer calibration, as illustrated in Fig. \ref{fig: field_xy_error}. Even minor biases in calibration parameters result in significant divergence during straight trajectory sections. Among the benchmark methods, \magfactor exhibited partial improvement over the raw measurements but still suffered from noticeable drift, leading to large deviations from the ground truth trajectory. Similarly, the Ellipsoid Fit method introduced inaccuracies that became evident during turns, causing the trajectory to diverge further. In contrast, the \magyc methods closely followed the ground truth trajectory, with minimal error accumulation, primarily occurring during turns and at significantly smaller magnitudes compared to the benchmark methods.

Using the EXP2-24 dataset, where the platform employed an active tilting actuator to dynamically track the seafloor, further evaluations of \ac{dr} performance revealed similar trends. As shown in Fig. \ref{fig: field_results_exp2_24}, even though the baseline error was smaller compared to EXP2-14, the \magyc methods still outperformed the benchmarks, reducing position errors to under \SI{4.68}{\meter} over a total \ac{dt} of \SI{1638}{\meter}. Benchmark methods like Ellipsoid Fit and \twostep converged but still fell short of the \magyc methods' accuracy. Notably, \magfactor demonstrated limited improvement over the non-calibrated data, yielding errors an order of magnitude worse than those achieved by the \magyc methods.

These results emphasize the substantial gains in position estimation accuracy provided by the \magyc methods, showcasing their robustness and effectiveness under diverse and challenging conditions in underwater navigation scenarios.

%%%%%%%%%%%%%%%%%%%%%%%%%%%%%%%%%%%%%%%%%%%%%%%%%%%%%%%%%%%%%%%%%%%%%%%%%%%%%%%%
% Conclusions
\section{Conclusions}
\label{sec:conclusion}

%%%%%%%%%%%%%%%%%%%%%%%%%%%%%%%%%%%%%%%%%%%%%%%%%%%%%%%%%%%%%%%%%%%%%%%%%%%%%%%%
% Figure: Field results EXP2-24
% Description: xy trajectory errors
%

\noindent The MAgnetometer and GYroscope Calibration (\magyc) methods—\magycbfg and \magycifg—demonstrated significant improvements in the performance of \ac{ahrs} sensors across both simulated and field scenarios. The results highlight that the \magyc methods consistently outperform or are comparable to previously proposed methods such as \twostep, Ellipsoid Fit, and \magfactor. These benchmark methods often rely on precise knowledge of the local magnetic field magnitude and system attitude or require a wide angular range for convergence, which limits their applicability to full-scale vehicles like \acp{rov}.

Numerical simulations showed that the \magyc methods excel in various scenarios, not only surpassing benchmark methods in their ideal conditions but also maintaining robustness under constrained angular movements. Additionally, field evaluation results underscore the substantial enhancements in position estimation accuracy achieved by the \magyc methods, demonstrating their effectiveness in diverse and challenging underwater navigation conditions. These methods achieved substantial reductions in heading errors compared to previously reported techniques, thereby enhancing underwater vehicle navigation, particularly in standard oceanographic surveys, and reaffirming their practical applicability.

In conclusion, the \magyc methods proposed in this study offer a viable and effective solution for calibrating magnetometers and gyroscopes for attitude estimation. These findings are expected to benefit the development of low-cost navigation systems and enhance the performance of ground, marine, and aerial vehicles in real-world applications.

%%%%%%%%%%%%%%%%%%%%%%%%%%%%%%%%%%%%%%%%%%%%%%%%%%%%%%%%%%%%%%%%%%%%%%%%%%%%%%%%
% Acknowledgement
\section*{Acknowledgments}

\noindent The field experimental data used in this study were collected in oceanographic surveys conducted by MBARI, led by Chief Scientist Dr. David Caress. Additionally, we acknowledge Dinale \cite{Dinale2013} and Bazhin et al. \cite{Bazhin2022} for providing open-source code for the implementation of the \twostep \cite{Alonso2002b} and Ellipsoid Fit methods, respectively.

%%%%%%%%%%%%%%%%%%%%%%%%%%%%%%%%%%%%%%%%%%%%%%%%%%%%%%%%%%%%%%%%%%%%%%%%%%%%%%%%
% References
\bibliography{refs/IEEEabrv, refs/MagCal}
\bibliographystyle{IEEEtran}

\end{document}